# The Judge Variable
## *Challenging Judge-Agnostic Legal Judgment Prediction*

**Guillaume ZAMBRANO,**
*Associate Professor, Faculty of Law, Nîmes University (France)*
ORCiD ID: https://orcid.org/0000-0002-6661-6612

**Abstract.** This study examines the role of human judges in legal decision-making by using machine learning to predict child physical custody outcomes in French appellate courts. Building on the legal realism-formalism debate, we test whether individual judges' decision-making patterns significantly influence case outcomes, challenging the assumption that judges are neutral variables that apply the law uniformly. To ensure compliance with French privacy laws, we implement a strict pseudonymization process. Our analysis uses 18,937 living arrangements rulings extracted from 10,306 cases. We compare models trained on individual judges' past rulings (specialist models) with a judge-agnostic model trained on aggregated data (generalist models). The prediction pipeline is a hybrid approach combining large language models (LLMs) for structured feature extraction and ML models for outcome prediction (RF, XGB and SVC). Our results show that specialist models consistently achieve higher predictive accuracy than the general model, with top-performing models reaching F1 scores as high as 92.85%, compared to the generalist model's 82.63% trained on 20x to 100x more samples. Specialist models capture stable individual patterns that are not transferable to other judges. In-Domain and Cross-Domain validity tests provide empirical support for legal realism, demonstrating that judicial identity plays a measurable role in legal outcomes. All data and code used will be made available.

## 1. Introduction

### *a Context*

Lady Justice holding her scales is a symbol of the fairness of the judiciary obeying the maxim "like cases should be treated alike." But is that really the case? As observed by Wu in a 2025 paper titled Reducing Judicial Inconsistency through AI, “inconsistencies not only impact the perceived fairness of the judicial system but also erode public trust in it.” [1] What would happen if we were to give exactly the same case to two different judges? Would their rulings be identical or divergent?

If we are to believe the supporters of Legal Idealism, the judge is - because he should be - merely as a "mouthpiece of the law" (Montesquieu, L’Esprit Des Lois, 1764). The law is an objective system that is speaking through the judge acting as a neutral processor. Judicial subjectivity, in this framework, is a mortal sin threatening the very integrity of the Rule of Law. For Idealists, legal reasoning is a deterministic process, akin to a mathematical equation. Legal rules, when applied to specific facts, yield predictable and consistent outcomes through formal, rule-based logic.

On the other side of the argument, if we are to listen to supporters of Legal Realism, “Law is what the judge had for breakfast”[1], suggesting that judicial decisions are unavoidably shaped by personal factors. Judges, as human agents of the Law, are not strictly bound by the formal

1 Jerome FRANK, Law and the Modern Mind, 1930

rules they are supposed to enforce, because they have an "open-texture"[2]. The application of abstract norms to concrete cases inherently discretionary judgment. Rules are not autonomous. They need a human to decide to what and when to apply them. Law is not purely objective knowledge, but also a matter of subjective choice and preference.

For sure, the debate could be put to rest by a double-blind experiment where we can control for the "Human Variable" in the Law. We could simply test judges by giving them the exact same case, the exact same files containing the same information, in a controlled environment, and compare their answers. Unfortunately, recruiting judges to subject themselves to such an experiment has proven impractical. Judges tend to be offended by such suggestions of inconsistency and subjectivity in rulings. And recreating a realistic a judicial environment in a controlled set-up is quite out of reach. Law is not yet an experimental science.

However, we could settle for the next best thing: simulating judges with Machine Learning models that will emulate their decisional profile based on past decisions, and compare the outcome of different models on the same case. This is the objective of the present study. We train specialist models on collected jurisprudence from identified judges, and train a generalist judge-agnostic model with similar cases from many judges.

### *b Research Hypotheses*

Our study uses 18,937 child custody rulings from French appellate courts to train two competing ML models: specialist models, trained only on cases decided by a single identified judge, and generalist models, trained on aggregate data of thousands of judges as if all judges were interchangeable. We then compare the results of all the models, on their own test set (in-domain validity), and all other test sets (cross-domain generalization). This methodological framework allows us to formulate testable hypotheses.

In a Legalist theoretical framework, knowledge of case facts and applicable legal rules should be enough to accurately predict the outcome of the case. The Legal Idealist Hypothesis predicts no significant difference between specialist models and generalist models. The law, being the same for everyone, a generalist model should yield consistent predictions regardless of the judge. If Legal Idealism accurately describes judicial decision-making, we should observe consistent prediction accuracy regardless of which judge decided the case. The generalist model would perform equally well across all judges, with predictive features maintaining consistent importance across different judicial officers. Furthermore, we would expect minimal systematic patterns in how judges deviate from the generic model, suggesting that objective application of legal rules predominates over individual judicial characteristics.

In contrast, the Legal Realist hypothesis expects superior performance from specialist models and a lack of generalization; specialist models should not transfer well across judges, exposing judge-specific "signatures" in decision-making. In a Realist theoretical framework, all things being equal, the same case ruled by two different judges can receive a different outcome. Realists postulate that outcome is judge-dependent. If the realist paradigm is true, we anticipate observing significant variations in prediction accuracy across different judges, with judge-specific models demonstrating superior performance compared to a judge-agnostic generic model. We would also expect different predictive features to carry varying importance depending on the judge, and systematic patterns should emerge in how specific judges deviate from the general jurisprudential norm. These observations would support the Realist contention that judges subjectivity substantially influences case outcomes beyond the objective facts presented.

With this simple experimental design we can test the reality of the "judge effect" on judicial outcome, and decide if the scales of Lady Justice are really untouched by the hands of the judges holding them.

## 2. Dataset

Our research strategically targets a legal domain where judges exercise maximal discretion with minimal statutory guidance. In the absence of explicit legal rules, judges must display

2 H.L.A HART, The Concept of Law, 1961

judicial discretion. To this purpose, we selected decisions by French Courts of Appeal concerning child living arrangements following parental separation (Child Physical Custody). The French legal framework establishes a broad decisional space for judges. The Civil Code (Article 373-2-9) merely outlines possible residential arrangements, stating that residence "*may be fixed alternately at each parent's home or at one of them*" without establishing presumptions or preferences among these options. Article 373-2-11 enumerates considerations judges may evaluate: previous arrangements, children's expressed preferences, parental capacities, expert evaluations, social inquiries, and inter-parental violence. These factors remain purely indicative. The legislative framework omits any prescribed methodology for weighing or prioritizing these elements. This absence of binding criteria or hierarchical factor analysis creates the precise conditions where judicial individuality might significantly influence outcomes. In contrast, U.S. jurisdictions operate under substantially more structured guidance. Unlike the French system, most U.S. states have enacted detailed statutory schemes that not only enumerate specific factors courts must consider but often provide explicit weighting instructions, presumptions, and procedural requirements governing judicial decision-making in custody matters. For instance, Minnesota's custody statute lists twelve specific factors that courts must evaluate when determining a child's best interests. The statute explicitly prohibits courts from using any single factor as determinative and requires judges to make "detailed findings on each of the factors and explain how the factors led to its conclusions and to the determination of custody and parenting time."[3]

The second reason for selecting Child Physical Custody cases is the quantity. Disputes on child living arrangements are very frequent, and generate abundant rulings, which we need to reconstitute significant temporal series. Child residence determinations represent a particularly consequential and high-conflict domain, with approximately 100,000 cases adjudicated annually in France [2]. The stakes for families are profound, affecting parent-child relationships, residential stability, educational continuity, and psychological well-being. Recent research [3] documents widespread perceptions of unfairness and procedural opacity among post-divorce families, contributing to systemic distrust of family court processes. Enhanced predictive capabilities could help demystify judicial reasoning patterns, making implicit decisional frameworks more explicit and potentially reducing perceptions of arbitrariness that undermine confidence in judicial outcomes.

Our dataset comprises judicial decisions obtained from the comprehensive JURICA database (JURIsprudence of Courts of Appeal), accessed through our university's institutional subscription. Since May 2020, decisions of Appellate Courts (civil law matters) have become freely available through the JUDILIBRE API, maintained by the French Court of Cassation and the Direction de l'Information Légale et Administrative (DILA). We collected 10,528 decisions spanning from 2007 to 2017, encompassing rulings from all 36 French Courts of Appeal. The **twenty-nine Metropolitan courts of appeal** are : Agen, Aix-en-Provence, Amiens, Angers, Besançon, Bordeaux, Bourges, Caen, Chambéry, Colmar, Dijon, Douai, Grenoble, Limoges, Lyon, Metz, Montpellier, Nancy, Nîmes, Orléans, Paris, Pau, Poitiers, Reims, Rennes, Riom, Rouen, Toulouse, Versailles. The **seven Overseas courts of appeal** are : Basse-Terre (Guadeloupe), Cayenne (French Guiana), Fort-de-France (Martinique), Nouméa (New Caledonia), Papeete (French Polynesia), Saint-Denis (Réunion), Bastia (Corsica).

A critical data limitation warrants a caveat: exhaustiveness of the JURICA database cannot be guaranteed. Indeed, closer inspection reveals that some training sets suffered from missing data. However, the large scale of our dataset provides a robust statistical buffer against the impact of missing entries. Crucially, we verified uninterrupted temporal sequences of decisions for the majority of judges, strongly suggesting that data absence is neither ubiquitous nor systematically biased. This imperfect yet viable scenario gives us confidence that the results presented remain fundamentally robust, even if some noise was introduced into the signal.

In this dataset, each document corresponds to a single case involving two parents engaged in a custody dispute. However, the number of children involved varies from case to case. Each child within a case may be subject to a separate custody determination, meaning that a single document (case) may contain multiple rulings. As a result, the total number of documents (cases) is not equal to the total number of rulings. The original dataset included 10,528 cases

3 Minn. Stat. § 518.17: <https://www.revisor.mn.gov/statutes/cite/518.17>

(single documents) corresponding to 19,903 child specific rulings. We excluded 966 rulings, with rarer outcome label, that did not fit in the simpler tri-category (mother, father, shared), and obtained our final dataset of 18937 rulings.

**Table 1. An Imbalanced Dataset.** *Class Distribution per Jurisdiction.*

| Group | Total | (%) overall | Class1 \| father (N) | Class1 \| father (%) | Class0 \| mother (N) | Class0 \| mother (%) | Class2 \| shared (N) | Class2 \| shared (%) |
|---|---|---|---|---|---|---|---|---|
| TOTAL | 18,937 | 100.0 % | 4,216 | 22.3 % | 12,741 | 67.3 % | 1,980 | 10.5 % |
| **METROPOLITAN COURTS** | | | | | | | | |
| Agen | 223 | 1.2 % | 66 | 29.6 % | 135 | 60.5 % | 22 | 9.9 % |
| Aix | 635 | 3.4 % | 124 | 19.5 % | 450 | 70.9 % | 61 | 9.6 % |
| Amiens | 269 | 1.4 % | 77 | 28.6 % | 171 | 63.6 % | 21 | 7.8 % |
| Angers | 268 | 1.4 % | 66 | 24.6 % | 176 | 65.7 % | 26 | 9.7 % |
| Besancon | 598 | 3.2 % | 164 | 27.4 % | 385 | 64.4 % | 49 | 8.2 % |
| **Bordeaux** | **1,255** | **6.6 %** | **315** | **25.1 %** | **798** | **63.6 %** | **142** | **11.3 %** |
| Bourges | 235 | 1.2 % | 59 | 25.1 % | 138 | 58.7 % | 38 | 16.2 % |
| **Caen** | **1,533** | **8.1 %** | **391** | **25.5 %** | **960** | **62.6 %** | **182** | **11.9 %** |
| Chambery | 518 | 2.7 % | 114 | 22.0 % | 336 | 64.9 % | 68 | 13.1 % |
| Colmar | 542 | 2.9 % | 100 | 18.5 % | 390 | 72.0 % | 52 | 9.6 % |
| Dijon | 655 | 3.5 % | 133 | 20.3 % | 464 | 70.8 % | 58 | 8.9 % |
| Douai | 751 | 4.0 % | 152 | 20.2 % | 548 | 73.0 % | 51 | 6.8 % |
| Grenoble | 904 | 4.8 % | 207 | 22.9 % | 600 | 66.4 % | 97 | 10.7 % |
| Limoges | 452 | 2.4 % | 106 | 23.5 % | 280 | 61.9 % | 66 | 14.6 % |
| **Lyon** | **1,613** | **8.5 %** | **335** | **20.8 %** | **1,132** | **70.2 %** | **146** | **9.1 %** |
| Metz | 444 | 2.3 % | 105 | 23.6 % | 300 | 67.6 % | 39 | 8.8 % |
| Montpellier | 169 | 0.9 % | 35 | 20.7 % | 125 | 74.0 % | 9 | 5.3 % |
| **Nancy** | **1,072** | **5.7 %** | **257** | **24.0 %** | **706** | **65.9 %** | **109** | **10.2 %** |
| Nimes | 419 | 2.2 % | 93 | 22.2 % | 291 | 69.5 % | 35 | 8.4 % |
| Orleans | 132 | 0.7 % | 29 | 22.0 % | 95 | 72.0 % | 8 | 6.1 % |
| **Paris** | **1,620** | **8.6 %** | **264** | **16.3 %** | **1,181** | **72.9 %** | **175** | **10.8 %** |
| Pau | 173 | 0.9 % | 26 | 15.0 % | 113 | 65.3 % | 34 | 19.7 % |
| Poitiers | 399 | 2.1 % | 98 | 24.6 % | 266 | 66.7 % | 35 | 8.8 % |
| Reims | 256 | 1.4 % | 66 | 25.8 % | 160 | 62.5 % | 30 | 11.7 % |
| Rennes | 714 | 3.8 % | 184 | 25.8 % | 460 | 64.4 % | 70 | 9.8 % |
| Riom | 63 | 0.3 % | 13 | 20.6 % | 44 | 69.8 % | 6 | 9.5 % |
| Rouen | 712 | 3.8 % | 151 | 21.2 % | 494 | 69.4 % | 67 | 9.4 % |
| Toulouse | 313 | 1.7 % | 67 | 21.4 % | 207 | 66.1 % | 39 | 12.5 % |
| **Versailles** | **1,309** | **6.9 %** | **260** | **19.9 %** | **897** | **68.5 %** | **152** | **11.6 %** |
| **OVERSEAS COURTS** | | | | | | | | |
| BasseTerre | 62 | 0.3 % | 18 | 29.0 % | 34 | 54.8 % | 10 | 16.1 % |
| Bastia | 120 | 0.6 % | 19 | 15.8 % | 90 | 75.0 % | 11 | 9.2 % |
| Cayenne | 12 | 0.1 % | 0 | 0.0 % | 11 | 91.7 % | 1 | 8.3 % |
| FortdeFrance | 28 | 0.1 % | 6 | 21.4 % | 20 | 71.4 % | 2 | 7.1 % |
| Noumea | 67 | 0.4 % | 15 | 22.4 % | 47 | 70.1 % | 5 | 7.5 % |
| Papeete | 30 | 0.2 % | 10 | 33.3 % | 17 | 56.7 % | 3 | 10.0 % |
| Saint-denis | 372 | 2.0 % | 91 | 24.5 % | 220 | 59.1 % | 61 | 16.4 % |

The data reveal a small number of metropolitan hubs handling a disproportionate share of the caseload. The top six cities representing more than a thousand samples each (Paris, Versailles, Nancy, Lyon, Bordeaux and Caen) account for 44.36% of all rulings (8402 rulings). Smaller cities and overseas territories (e.g., Papeete, Noumea, Cayenne) represent a minimal fraction of the dataset.

## 3. Structured Features Extraction With LLM

The study relies on three different features that need to be extracted separately using Llama 3.3 (70B) : *judge pseudonymized identity* (a), *case description features* (b), and *case outcome* (c). Only case outcomes were validated against a human expert manually annotated gold standard of 500 rulings.

### *a Judge Pseudonymized Identity*

French criminal law establishes a felony prohibition against judge profiling. Violations carry criminal penalties including imprisonment for up to five years and fines of up to 300,000 euros (French Penal Code, Article 226-18). The personal data (names) of magistrates and court clerks may not be reused for the purpose or effect of evaluating, analyzing, comparing, or predicting their actual or supposed professional practices (Code de l'organisation judiciaire, Article L111-13, since March 23, 2019). To comply with French criminal law, our pipeline is divided into two clear and separate phases : identity protection and outcome extraction. We splitted the original documents and compartmentalized sections containing judges identity and sections containing the outcome of the case. These sections were treated separately, ensuring that at no point the real identities of judges are linked to outcome data. This two-step workflow ensures that real identities are never connected to outcome data. In the first stage we extract the names of the judges with the objective of pseudonymization. We implemented a dictionary-based pseudonymization framework that systematically replaces identifying information with 7 letters words from the french dictionary. In the second stage, we extract the outcome of the case and link it with the pseudonymized identity of the judge, using the name of the file as a neutral key.

In order to train our specialist and generalist models, we partitioned the dataset based on judicial identity. This partitioning enables a comparative analysis between models trained on individual judges and those trained on aggregated judicial data. **For the specialist models**, we identified the 13 most represented presiding judges (judge1) each having ruled on more than 300 individual children, representing collectively 37% of the dataset. These rulings were divided into 13 buckets dedicated to the training of 13 specialist models. In the end, this bucketing results in a pretty small number of sampls for training, considering we have three very imbalanced classes (class 2 is roughly 10%) and we have to take away 20% of the 300 samples for testing. **For the generalist models**, we aggregated the remaining 11975 rulings (63%) into the judge-agnostic "generic" subset.

**Table 2. Judge Buckets.**

| Judge | City | Begin | End | Cases | Rulings |
|---|---|---|---|---|---|
| Anatole | Nancy | 2007 | 2015 | 457 | 824 |
| Babiche | Grenoble | 2008 | 2016 | 427 | 769 |
| Cabeche | Versailles | 2009 | 2017 | 395 | 731 |
| Dacrons | Caen | 2011 | 2015 | 348 | 635 |
| Echevin | Caen/Lyon | 2007 | 2017 | 309 | 590 |
| Faubers | Besançon | 2008 | 2017 | 299 | 549 |
| Gargote | Lyon | 2011 | 2015 | 295 | 572 |
| Hauynes | Bordeaux | 2011 | 2015 | 279 | 497 |
| Inconel | Caen | 2014 | 2017 | 229 | 408 |
| Jobelin | Chambery | 2014 | 2017 | 225 | 408 |
| Kochias | Saint Denis | 2008 | 2017 | 216 | 360 |
| Labourg | Pau/Paris | 2008 | 2017 | 174 | 308 |
| Orillon | Lyon | 2008 | 2012 | 154 | 311 |
| Specialists | - | 2007 | 2017 | 3807 | 6962 |
| Generalist | - | 2001 | 2022 | 6499 | 11975 |
| Total | - | 2001 | 2022 | 10306 | 18937 |

The top three judges (Anatole, Babiche and Cabeche) handled 1,279 cases, accounting for 12.27% of the entire total of samples. Expanding to the top five judges, the total rises to 1,919 cases, representing 18.23% of all cases. This means that fewer than 0.39% of judges adjudicated nearly one-fifth of the total caseload. The concentration becomes even more pronounced when examining the top 10 judges, who accounted for 27.1% of all cases, with the next 10 judges handling an additional 20.5%.Together, the top 20 judges (representing just 1.5% of all judges) were responsible for nearly half (47.6%) of the entire caseload.

**Table 3. Imbalanced Buckets.** *Class Distribution per Judge.*

| Group | Total | % of overall | Class1 \| father (N) | Class1 \| father (%) | Class0 \| mother (N) | Class0 \| mother (%) | Class2 \| shared (N) | Class2 \| shared (%) |
|---|---|---|---|---|---|---|---|---|
| anatole | 824 | 4.4 % | 192 | 23.3 % | 555 | 67.4 % | 77 | 9.3 % |
| babiche | 769 | 4.1 % | 180 | 23.4 % | 507 | 65.9 % | 82 | 10.7 % |
| cabeche | 731 | 3.9 % | 161 | 22.0 % | 462 | 63.2 % | 108 | 14.8 % |
| dacrons | 635 | 3.4 % | 152 | 23.9 % | 412 | 64.9 % | 71 | 11.2 % |
| echevin | 590 | 3.1 % | 156 | 26.4 % | 365 | 61.9 % | 69 | 11.7 % |
| faubers | 549 | 2.9 % | 151 | 27.5 % | 349 | 63.6 % | 49 | 8.9 % |
| gargote | 572 | 3.0 % | 102 | 17.8 % | 430 | 75.2 % | 40 | 7.0 % |
| hauynes | 497 | 2.6 % | 126 | 25.4 % | 318 | 64.0 % | 53 | 10.7 % |
| inconel | 408 | 2.2 % | 103 | 25.2 % | 256 | 62.7 % | 49 | 12.0 % |
| jobelin | 408 | 2.2 % | 98 | 24.0 % | 259 | 63.5 % | 51 | 12.5 % |
| kochias | 360 | 1.9 % | 91 | 25.3 % | 209 | 58.1 % | 60 | 16.7 % |
| labourg | 308 | 1.6 % | 42 | 13.6 % | 229 | 74.4 % | 37 | 12.0 % |
| orillon | 311 | 1.6 % | 72 | 23.2 % | 203 | 65.3 % | 36 | 11.6 % |
| **generic** | 11,975 | 63.2 % | 2,590 | 21.6 % | 8,187 | 68.4 % | 1,198 | 10.0 % |
| TOTAL | 18,937 | 100 % | 4,216 | 22.3 % | 12,741 | 67.3 % | 1,980 | 10.5 % |

*b   Facts of the case*

In legal artificial intelligence, case outcome prediction falls under two distinct paradigms: feature-based (structured) and text-based (unstructured) prediction models. Contemporary Legal Judgment Prediction (LJP) research predominantly adopts this text-based paradigm. However, the earliest computational models in legal AI were symbolic feature-based systems [4, 5, 6]. Despite their strengths, symbolic models have historically been constrained by the labor-intensive nature of feature annotation. As Ringger et al. observed [7], annotation is often the most resource-intensive component of corpus development. This annotation bottleneck has been cited repeatedly in the literature as a key limitation to scaling structured approaches [8,9].

In this study, we combine feature-based ML methods with LLM data extraction . We use LLMs to create structured feature representations from case text, then apply standard ML models to predict outcomes based on extracted features. LLMs have demonstrated zero-shot capabilities for complex information extraction tasks, making them well-suited to automate the identification of legal features from unstructured texts as illustrated by recent studies [10, 11, 12]. Our pipeline uses Meta's Llama 3.3 70B to extract structured features. All features are categorical.

The Petitions Features cluster (Cluster 1) tests our hypothesis that parental petitions and child expressed preferences significantly influence judicial outcomes. This hypothesis is grounded in Civil Law. Judges must rule solely on what is requested by the parties (Code de Procédure Civile, Article 5). The passive role of judges in shaping litigation is a legacy from Roman law principle "nemo judex sine actore" ("no judge without a plaintiff"). Courts are prohibited from ruling beyond or outside the limits of the dispute as defined by the parties' claims, a restriction known as the prohibition against ruling "infra petita" or "extra petita."

The Behavioral Features cluster (cluster 2) comprises psychosocial risk factors, which include indicators such as parental addiction, abuse directed at the child or the other parent, neglect or diagnosed psychological disorders. These factors would reflect the court's overarching duty to prioritize the child's safety and psychological well-being embedded in the "best interests of the child" standard, which prioritizes child protection over parental rights when the two come into conflict (United Nations Convention on the Rights of the Child, Article 3). Among the 17,708 fathers in our dataset (excluding 180 rulings with missing values), abuse by the father was reported in 1,229 rulings, representing 6.9% of all fathers. In contrast, among the 18,550 mothers (excluding 212 missing), abuse by the mother was reported in only 387 rulings, or 2.1%.

The Practical Features cluster (cluster 3) acknowledges that courts often favor arrangements that minimize disruption to the child's social and educational environment. These logistical factors are often explicitly discussed in judicial opinions.

**Table 4. Facts of the case.** *Extracted features. All features are categorical.*

| IDX | Features (Cluster 1) | Values |
|---|---|---|
| 12 | child_expressed_conflict | Mother, father, both, none |
| 13 | child_expressed_living_arrangement_preference | Mother, father, shared, none |
| 14 | father_request_regarding_living_arrangements | None, Sole, Shared |
| 16 | mother_request_regarding_living_arrangements | None, Sole, Shared |
| **IDX** | **Features (Cluster 2)** | **Values** |
| 18 | father_parental_fitness | boolean |
| 19 | mother_parental_fitness | boolean |
| 20 | father_has_history_of_abuse_against_child | boolean |
| 21 | mother_has_history_of_abuse_against_child | boolean |
| 22 | father_has_history_of_abuse_against_mother | boolean |
| 23 | mother_has_history_of_abuse_against_father | boolean |
| 24 | father_has_history_of_neglect | boolean |
| 25 | mother_has_history_of_neglect | boolean |
| 26 | father_has_psych_issues | boolean |
| 27 | mother_has_psych_issues | boolean |
| 28 | father_has_addiction_issues | boolean |
| 29 | mother_has_addiction_issues | boolean |
| 30 | father_is_invested_with_child | boolean |
| 31 | mother_is_invested_with_child | boolean |

| IDX | Features (Cluster 3) | Values |
|---|---|---|
| 32 | father_employment_status | boolean |
| 34 | mother_employment_status | boolean |
| 36 | father_work_availability | boolean |
| 37 | mother_work_availability | boolean |
| 38 | father_housing_status | boolean |
| 39 | mother_housing_status | boolean |
| 40 | Parent Proximity | boolean |
| 41 | father_lives_near_school | boolean |
| 42 | mother_lives_near_school | boolean |
| 43 | father_receives_social_aid | boolean |
| 44 | mother_receives_social_aid | boolean |
| 56 | mother_benefited_legal_aid | boolean |
| 57 | father_benefited_legal_aid | boolean |

### *c Case Outcome Extraction*

In this section, we evaluate the accuracy of a large language model (LLaMA 3.3 70B) in extracting structured child physical custody outcome. For baseline, we compared Llama with human annotators, specifically undergraduate students (third-year junior-level law students). Both human annotators and LLM were evaluated against a Gold Standard established by the author. We chose not to use inter-annotator agreement as a proxy for ground truth because legal outcomes are objective and not open to interpretation. Our methodology is aligned with best practices in the legal NLP field [13].

Nonetheless, these documents are quite complex to understand, even for legally trained humans. Student annotators were compensated with academic credits and received detailed written and verbal guidelines. Annotations were conducted over a two-week period, with daily supervision and review by the lead author. Despite close monitoring, the annotations produced by students were frequently inconsistent and error-prone, particularly in complex cases.

One major source of confusion was the instruction to extract the outcome of the trial court, due to the frequent presence of multiple and conflicting decisions. Another challenge was resolving references to the factual and procedural history scattered across the decision. Students found it particularly difficult to maintain consistency in tracking all entities, especially when the judgment covered multiple children. Performance varied sharply between trial and appellate court decisions. At the trial level, inter-annotator agreement reached only moderate levels ($\kappa = 0.6376$), while appeal court annotations achieved near-perfect agreement ($\kappa = 0.8869$). This discrepancy reflects differences in legal drafting. Trial court rulings often include a full procedural history, sometimes referencing multiple earlier decisions, including instances where earlier judgments were reversed or modified. This complexity makes it difficult for annotators to identify which trial decision is the operative one prior to appeal.

Documents are also quite long. Key informations are disseminated in the text or never explicited. Facts, legal reasoning, and judges opinions are woven together in a single narrative. There is no standardized format, styles can vary. Outcomes can appear in scattered clauses, conditional statements, or indirect references to prior rulings that must be retrieved or inferred. This unstructured format makes sentence-level extraction unreliable and calls for document-level understanding.

Another difficulty is the high degree of lexical and syntactic variation. A single legal concept can be written in many different ways. For example, child support may appear as "pension alimentaire", "contribution à l'entretien", or described through legal obligations like "mise à la charge du père". Similarly, visitation rights are often vaguely expressed by reference to customs of the courts, e.g., "suivant les modalités habituelles". These variations favor models able to generalize semantically and understand legal context.

We evaluated LLM Data extraction across four distinct runs with different prompts. Prompt#1 (1344 characters – 337 tokens) is the minimal prompt. Prompt#2 (2642 characters – 624 tokens), Prompt#3 (2214 characters – 553 tokens) and Prompt#4 (6061 characters – 1403 tokens) are increasingly long and detailed variations. We only present Prompt#1 that was used for extraction. Prompts #2, #3 and #4 were only intended for investigation.

```
SYSTEM_ROLE: LEGAL_DATA_EXTRACTION_AGENT
DOMAIN: French legal rulings from trial court ("juge aux affaires familiales" - JAF) and appeal
court (cour d'appel - CAPP) on child physical custody, visitation rights and child support
GOAL: Extract structured JSON data based on information found in legal rulings.
CONSTRAINTS:
- Extract child-specific data for every child mentioned in the dispute
- Answer all fields
- Output format: strict JSON schema
- JSON only – no extra text, no markdown
SCHEMA_START =>
<JSON_SCHEMA>
case_metadata:
  appeal_decision_date: yyyy-mm-dd
  appeal_case_id: numero_RG
  appeal_court_city: string
child_specific_outcome[]:
  child_first_name: string
  just_before_appeal_review_child_living_with: mother|father|joint|other
  just_before_appeal_review_visitation_rights_granted_to: mother|father|both|other
  just_before_appeal_review_child_support_per_child_per_month_is_equal_to: int
  just_before_appeal_review_child_support_paid_by: mother|father|nopayer
  after_appeal_review_child_living_with: mother|father|joint|other
  after_appeal_review_visitation_rights_granted_to: mother|father|both|other
  after_appeal_review_child_support_amount_per_child_per_month_is_equal_to: int
  afer_appeal_review_child_support_paid_by: mother|father|nopayer
</JSON_SCHEMA>
<= SCHEMA_END
EXECUTE_EXTRACTION_MODE: STRICT
```

**Figure 1. Case Outcome Extraction Prompt.** *Prompt#1*

**Table 5. Extraction Performance: Child Physical Custody (Trial Court Outcome).**

| Run# | Cohen k | TP | Precision | Recall | F1 |
|---|---|---|---|---|---|
| | | Father Custody (trial_court) | | | |
| human | 0.64 | 66/110 | 0,96 | 0,60 | 0,74 |
| Run#1 | 0,6922 | 122/157 | 0,85 | 0,78 | 0,81 |
| Run#2 | 0,8249 | 132/157 | 0,94 | 0,84 | 0,89 |
| Run#3 | 0,7863 | 122/157 | 0,97 | 0,78 | 0,86 |
| Run#4 | 0,8551 | 144/158 | 0,94 | 0,91 | 0,92 |
| | | Joint custody (trial court) | | | |
| human | 0.64 | 63/70 | 0,67 | 0,90 | 0,77 |
| Run#1 | 0,6922 | 51/82 | 1,00 | 0,62 | 0,77 |
| Run#2 | 0,8249 | 73/82 | 0,96 | 0,89 | 0,92 |
| Run#3 | 0,7863 | 70/82 | 0,89 | 0,85 | 0,87 |
| Run#4 | 0,8551 | 70/82 | 0,95 | 0,85 | 0,90 |
| | | Mother custody (trial court) | | | |
| human | 0.64 | 151/179 | 0,76 | 0,84 | 0,80 |
| Run#1 | 0,6922 | 234/251 | 0,80 | 0,93 | 0,86 |
| Run#2 | 0,8249 | 241/251 | 0,87 | 0,96 | 0,91 |
| Run#3 | 0,7863 | 243/251 | 0,83 | 0,97 | 0,90 |
| Run#4 | 0,8551 | 237/251 | 0,91 | 0,94 | 0,93 |

**Table 6. Extraction Performance: Child Physical Custody (Appeal Court Outcome).**

| Run# | Cohen k | TP | Precision | Recall | F1 |
|---|---|---|---|---|---|
| | | Father Custody (appeal_court) | | | |
| human | 0.89 | 83/91 | 0,90 | 0,91 | 0,91 |
| Run#1 | 0,8469 | 120/124 | 0,92 | 0,97 | 0,94 |
| Run#2 | 0,8504 | 120/124 | 0,92 | 0,97 | 0,94 |
| Run#3 | 0,8914 | 122/124 | 0,92 | 0,98 | 0,95 |
| Run#4 | 0,8715 | 115/125 | 0,90 | 0,92 | 0,91 |
| | | Joint Custody (appeal_court) | | | |
| human | 0.89 | 48/49 | 0,89 | 0,98 | 0,93 |
| Run#1 | 0,8469 | 56/65 | 0,95 | 0,86 | 0,90 |
| Run#2 | 0,8504 | 57/65 | 0,95 | 0,88 | 0,91 |
| Run#3 | 0,8914 | 63/65 | 0,95 | 0,97 | 0,96 |
| Run#4 | 0,8715 | 64/65 | 0,96 | 0,98 | 0,97 |
| | | Mother Custody (appeal_court) | | | |
| human | 0.89 | 189/201 | 0,96 | 0,94 | 0,95 |
| Run#1 | 0,8469 | 279/285 | 0,93 | 0,98 | 0,95 |
| Run#2 | 0,8504 | 279/285 | 0,93 | 0,98 | 0,95 |
| Run#3 | 0,8914 | 283/285 | 0,94 | 0,99 | 0,97 |
| Run#4 | 0,8715 | 274/285 | 0,94 | 0,96 | 0,95 |

Results show that LLM data extraction is comparable to human expert annotation and reliable enough. In our initial configuration (Prompt #1), we employed a minimal-prompt strategy. The prompt consisted of a bare JSON schema that defined only the expected output structure, without additional instructions, examples, or explanatory context. The total prompt length was 1,344 characters (337 tokens), following a minimalist prompt strategy [14,15]. The schema relied on closed ontologies (e.g., `father`, `mother`, `joint`), which constrained the model's responses to a predefined set of possible values. This design choice reduced ambiguity and minimized the risk of inconsistent or free-form outputs. As shown in the literature [16], schema-constrained prompting improves the reliability, interpretability, and utility of LLM outputs in classification and structured prediction tasks. By defining a fixed set of allowed values for each field, the prompt limited variation and helped control the model's output space. This prompt format thus mirrors best practices for improving consistency in zero-shot extraction settings.

When compared to human annotators, the LLM demonstrated higher overall accuracy, especially in the more challenging trial court cases. Human annotators frequently struggled to track multiple entities, resolve references to prior decisions, and disambiguate outcomes embedded in complex legal narratives. These findings show that large language models, when guided by minimal schema-based prompts, can already approximate or surpass human performance in structured legal information extraction. The high performance achieved with such minimal supervision highlights the scalability and practical value of LLMs in legal annotation tasks.

## 4. Literature Review

Extensive literature reviews of Legal Judgment Prediction have been published in 2019 by Ashley [17], in 2021 by Rosili, Zakaria & Hassan [18] and in 2022 by Feng, Li & Ng [19]. However, in this paper we would like to highlight the significant heterogeneity in datasets, task formulations, and evaluation metrics, coupled with frequent issues of data incompleteness. Thus, we adopt a dataset-centric approach to the literature, as each dataset introduces specific constraints and necessitates distinct methodological choices.

### *a United States Supreme Court (SCOTUS) datasets*

Predicting SCOTUS decisions is an old tradition in US Political Science. SCOTUS prediction is based almost exclusively on human expert-annotated datasets. As noted by Medvedeva et al. "The advantage of working with the SCOTUS database is that due to the attention it attracts, all trial data has systematically and manually been annotated with hundreds of variables by legal experts, shortly after the case has been tried." [20] However, for computational legal prediction models to demonstrate genuine analytical value in this domain, they must outperform strong baselines that can be achieved with naïve heuristics. Simply predicting that the petitioner will win yields nearly 68% accuracy, because the Court only grants certiorari (right to appeal) when there is a plausible basis for reversal [21,22]. Adopting a blanket "reverse" prediction strategy aligns with 63% case outcomes [23]. These heuristics highlight a structural bias in the dataset. The inherent selection bias in the Court's docket creates a statistical anomaly: most cases are non-random and selected for review by the Supreme Court. Consequently, any predictive model must outperform this strong baseline to demonstrate added value.

**Table 7. The SCOTUS datasets.** *Predicting United States Supreme Court.*

| Paper | Size | Model | Accuracy | F1 |
|---|---|---|---|---|
| [24] Ruger et al., 2004 | 696 | Classification Tree | 75.00 | - |
| [25] Katz et al., 2014 | 7,7k | Random Forest | 69.61 | 39.70 |
| [26] Sharma et al., 2014 | 7,7k | DNN | 70.4 | - |
| [27] Katz et al., 2017a | 28k | Random Forest | 70.02 | 69.00 |
| [28] Katz et al., 2017b | 450 | CrowdSourcing (FantasySCOTUS) | 80.8 | 85.2 |
| [29] Kaufman et al., 2019 | - | ADA Boosted Decision Tree | 74.04 | - |
| [30] Alghazzawi et al., 2022 | unclear | CNN+BiLSTM | 92.05 | 93 |

### b Other Supreme Courts around the world

Contrasting with SCOTUS predictive research, efforts on other supreme courts around the world have been focused on text-based features. From France to Brazil, scholars have deployed a range of models, from classic support vector machines to transformer-based architectures like BERT and GPT, to tackle the binary task of predicting whether a court will affirm or reverse a decision.

**Table 8. A Diverse World of Supreme Courts Datasets.**

| Paper | Dataset | Size | Balance | Model | F1 | Acc. |
|---|---|---|---|---|---|---|
| [31] Sulea, 2017a | Fra-CCass | 126k | 43 / 54 | SVM | 96.9 | 97.0 |
| [32] Sulea, 2017b | Fra-CCass | 126k | 43 / 54 | Ensemble | 98.6 | 98.6 |
| [33] AlMuslim, 2022 | Can-Appeal | 36k | 28 / 71 | BERT | 81.79 | - |
| [34] Strickson, 2020 | UK-SC | 5k | 50 / 50 | LR+TFIDF | 69.02 | 69.05 |
| [35] Niklaus, 2021 | Switz-SC | 85k | 23 / 77 | BERT | 68.5 | - |
| [36] Malik, 2021 | India-SC | 35k | 41 / 59 | XLNET+BiGRU | 77.79 | 77.78 |
| [37] Nigam, 2023 | India-SC | 35k | 41 / 59 | GPT3.5 | 73.98 | - |
| [38] Lage-Freitas, 2022 | Brazil-SC | 4k | 41 / 59 | XGBoost | 80.22 | 81.35 |

The LJP landscape remains notably fragmented and characterized by inconsistent standards. For instance, the exceptionally high predictive performance reported by Sulea et al. invites a critical examination of the underlying textual features driving these results. Such metrics strongly suggest the presence of strong outcome proxies embedded within the text input. In the context of French legal writing, it is very well known that the Cour de Cassation adheres to a highly formalized drafting style that varies systematically depending on the outcome of the case. For instance, in cases where the appeal is rejected ("rejet"), the decision typically introduces the plaintiff's argument with the phrase "alors selon le moyen que," signaling a negative resolution. In contrast, when the court overturns the lower ruling ("cassation"), the decision is introduced with "alors que," subtly but unmistakably indicating a favorable outcome. Such linguistic markers, along with changes in paragraph ordering and rhetorical structure, effectively render the outcome legible in the language of the ruling itself—well before any substantive legal reasoning is parsed. As a result, machine learning models trained on such texts may not be learning to predict judicial outcome in any meaningful sense but rather to detect stylistic and structural cues that correlate directly with the outcome.

### c European Court of Human Rights (ECHR) Datasets

The European Court of Human Rights (ECHR) has been the focus of attention for european researchers. Aletras et al. (2016) initially reported a significant 79% accuracy using a support vector machine (SVM) on a balanced dataset of 584 cases, a result that drew attention to the potential of machine learning in legal outcome prediction. However, this work faced immediate criticism, from Visentin, Nardotto, and O'Sullivan (2019) and Medvedeva, Vols, and Wieling (2019). Critics highlighted the small sample size and lack of reproducibility in Aletras' original work but also showed that, when applying similar methods to larger datasets, the resulting accuracies were consistently lower. In response to these issues, Chalkidis, Androutsopoulos, and Aletras (2019) used neural network models on an expanded dataset of 11,000 cases. These models achieved incremental accuracy improvements up to 82%, demonstrating the advantages of deep learning and larger, more representative datasets. More recent studies expanded the dataset to 20,000 cases and applied advanced transformer-based models but reported more modest results.

**Table 9. European Court of Human Rights datasets (ECHR)**

| Paper | Dataset | Size | Balance | Model | F1 (+) | F1 (-) | Accur. |
|---|---|---|---|---|---|---|---|
| [39] Aletras, 2016 | ECHR | 584 | 50/50 | SVM | - | | 79.0 |
| [40] Visentin, 2019 | ECHR-art3 | 584 | 50/50 | Ensemble | - | | 73.9 |
| [40] Visentin, 2019 | ECHR-art6 | 584 | 50/50 | Ensemble | - | | 86.1 |
| [40] Visentin, 2019 | ECHR-art8 | 584 | 50/50 | Ensemble | - | | 78.0 |
| [41] Medvedeva, 2018 | ECHR | 11k | 50/50 | SVM | 74.0 | | - |
| [42] Medvedeva, 2020 | ECHR-art2 | 11k | 50/50 | SVM | 72.0 | 70.0 | - |
| [42] Medvedeva, 2020 | ECHR-art3 | 11k | 50/50 | SVM | 80.0 | 79.0 | - |
| [42] Medvedeva, 2020 | ECHR-art6 | 11k | 50/50 | SVM | 80.0 | 82.0 | - |
| [42] Medvedeva, 2020 | ECHR-art8 | 11k | 50/50 | SVM | 69.0 | 72.0 | - |
| [42] Medvedeva, 2020 | ECHR-art10 | 11k | 50/50 | SVM | 63.0 | 65.0 | - |
| [43] Chalkidis, 2019 | ECHR | 11k | 50/50 | Bi-GRU | 79.5 | | - |
| [43] Chalkidis, 2019 | ECHR | 11k | 50/50 | HAN | 80.5 | | - |
| [43] Chalkidis, 2019 | ECHR | 11k | 50/50 | Hier-BERT | 82.0 | | - |
| [44] Xu, 2024 | ECHR | 11k | unclear | BERT | 69.3 | | - |
| [45] Cao, 2024 | ECHR | 20k | unclear | BERT | 71.15 | | - |

### *d Criminal Law Datasets*

In the domain of criminal law, the methodological and conceptual approach for LJP departs notably from ECHR or SCOTUS studies. In terms of quantity, Chinese researchers benefit from access to China Judgments Online (CJO), Peking University Law (PKULAW), and the datasets made available through the Challenge on AI and Law (CAIL). Xiao Chaojun et al. [46] proposed the CAIL2018 dataset containing 2.6 million criminal cases published by the Supreme People's Court of China. In terms of methodology, criminal LJP research is focused on prediction of an outcome triplet : legal basis for the charges (references of Law articles that are the normative statements defining the charges), the identification of charges (categorical class), and the sentence (ranging from not guilty to a certain duration in months). This framework has become the dominant paradigm in criminal LJP.

**Table 10. Criminal Law Datasets.**

| Paper | Country | Dataset | Size | Model family | Charges Accur. | Charges F1 | Sentence Accur. |
|---|---|---|---|---|---|---|---|
| [48] Luo, 2017 | China | CJO | 60k | AttentionNN | - | 95.42 | - |
| [49] Zhong, 2018 | China | CJO | 1007k | CNN-LSTM | 94.9 | 49.1 | 58.8 |
| [49] Zhong, 2018 | China | PKU | 177k | CNN-LSTM | 95.6 | 70.9 | 57.8 |
| [49] Zhong, 2018 | China | CAIL2018 | 113k | CNN-LSTM | 85.7 | 78.3 | 38.3 |
| [50] Hu, 2018 | China | CJO | 77k | HAN-LSTM | 93.4 | 64.9 | - |
| [50] Hu, 2018 | China | CJO | 192k | HAN-LSTM | 94.4 | 67.1 | - |
| [50] Hu, 2018 | China | CJO | 383k | HAN-LSTM | 95.8 | 73.1 | - |
| [51] Wei, 2019 | China | CAIL2018 | 204k | TextCNN | 86.23 | 77.50 | - |
| [52] Yang, 2019 | China | CAIL2018 | 101k | AttentionNN | 88.70 | 85.90 | 41.40 |
| [52] Yang, 2019 | China | CAIL2018 | 1588k | AttentionNN | 97.70 | 86.70 | 60.40 |
| [53] Xu, 2020 | China | CAIL2018 | 128k | GraphNN | 85.07 | 82.74 | 38.29 |
| [53] Xu, 2020 | China | CAIL2018 | 1746k | GraphNN | 96.45 | 85.35 | 59.66 |
| [54] Yue, 2021 | China | CAIL2018 | 134k | Bi-GRU | 89.92 | 86.96 | 41.65 |
| [54] Yue, 2021 | China | CAIL2018 | 1779k | Bi-GRU | 95.57 | 80.54 | 57.07 |
| [55] Dong, 2021 | China | CAIL2018 | 128k | Transformers | 89.13 | 87.94 | 43.77 |
| [55] Dong, 2021 | China | CAIL2018 | 1773k | Transformers | 97.93 | 91.94 | 64.71 |
| [56] Liu, 2023 | China | LAIC2021 | 98k | GraphNN | 97.54 | 94.64 | 47.63 |
| [57] Zhang, 2025 | China | CAIL2018 | 82k | BERT+LLM | 96.00 | 96.10 | 54.72 |
| [58] Peng, 2024 | Taiwan | No name | 218k | BERT | 98.95 | 98.85 | 80.93 |
| [59] Varga, 2021 | Slovak. | No name | 226k | CNN | 99.24 | 98.83 | - |

It should be noted that it is a bit odd to treat sentence as a single class containing "not guilty" and a range of numeric values. For instance, Peng [58] treats Sentence as a categorical class by creating three labels: majority class 0 "Acquittal to less than 4 months imprisonment" (57%), minority class 1 "4 months imprisonment to less than 6 months imprisonment" (27%), and minority class 2 "Over 6 months imprisonment" (16%). This means the models are treating as a single label, cases which resulted in acquittal and cases which resulted in imprisonment. It would seem more logical to predict a binary guilty / not guilty verdict first, and then predict the penalty quantum (sentence duration in months) if guilty. It may explain why the F1 score for predicting the sentence is very poor compared to the prediction of the legal basis in the high 90's.

### *e On judicial inconsistency*

The consistency of judicial decision-making, particularly how outcomes are influenced by the judge's identity, presents a major challenge for computational legal analysis. This area remains significantly underexplored, a fact noted by Wang Yuzhong et al. [60] who stated that "legal judgment consistency analysis over large-scale data remains to be explored." A 2023 survey by Wu Yifan illustrates the prevailing narrative within LJP research, centered on the development of AI as a tool to enhance judicial consistency and reduce outcome disparities [61]. This paradigm operates on the assumption that judicial discretion and individual variation are flaws to be engineered out of the system. The resulting technical objective is to create a "one-size-fits-all" model—a generalist system designed to produce a single, consistent outcome across all cases, effectively equating this statistical uniformity with the correct application of the law. In direct contrast, our research challenges this judge-agnostic premise. Instead of treating judicial identity as a noise or bug to be eliminated, we investigate it as a core, predictive variable in its own right.

A 2020 work by Medvedeva, Vols and Wielding [62] provided striking evidence for the "judge variable" by demonstrating that judicial outcomes could be predicted with significant accuracy based on extra-legal factors alone. In a starkly simple experiment, a model was trained using only the surnames of the judges presents in a case (even if this information is not necessarily known to the parties before hand). This minimalist approach achieved 65% accuracy, far exceeding a random guess and highlighting the predictive power of a judge's identity. As the authors noted, this shows that "the decision is influenced to a large extent by the judges in the Chamber". This led to the conclusion that even without knowing specific case details, "the identity of the judges is a useful predictor". However, the methodology treated judge identity as a single, collective feature within a pool of data. It demonstrated that the judiciary's composition matters, but did not delve into how the specific, replicable decision-making patterns of individual judges contribute to that effect.

A pivotal inspiration for the present study is the 2021 paper of Wang Yuzhong et al. [60], who developed the Legal Inconsistency Coefficient (LInCo) to measure outcome disparities. Their research aims to provide an empirical tool for assessing the principle of "equality before the law" by detecting and quantifying systemic variations in judgments. The LInCo framework is built on the concept of "virtual judges", i.e. machine learning models trained on specific subsets of legal data to simulate the decision-making patterns of a particular group. The core idea is that if different virtual judges, trained on data from different groups, consistently disagree on the outcome of the same case, it reflects a real-world inconsistency. We adopt the basic same methodology.

The authors first validate their metric through a carefully designed simulation study. This validation process begins with a single, unified set of cases. They then create multiple synthetic datasets by keeping the case facts identical across all sets but deliberately altering the sentence outcomes in each one. For example, one synthetic dataset might have sentences that are systematically increased by a certain factor, while another has them decreased. By training a separate virtual judge model on each of these artificially biased datasets, they create a controlled environment where the level of inconsistency is known beforehand. The LInCo score is then calculated by measuring the average disagreement between these models' predictions. The successful correlation between their known, injected bias and the resulting LInCo score serves to prove that the metric is a reliable instrument for detecting judicial inconsistency.

Having established the metric's validity, the LinCo is applied to a large, real-world dataset of Chinese criminal cases. For their analysis of regional inconsistency, they partition the entire corpus of cases based on the province where the judgment was rendered. A separate model, or "virtual judge," is then trained exclusively on the historical cases from each specific region (e.g., a Beijing model, a Shanghai model). These models learn the collective sentencing patterns and implicit biases of their respective regional judiciaries. The LInCo is then calculated by feeding a common set of test cases to all of these region-specific models and measuring the average difference in their predicted sentences. Their findings reveal a significant regional inconsistency, thereby providing empirical evidence that a case's outcome can be influenced by the jurisdiction in which it is tried.

This framework provides a direct methodological antecedent to our own research, yet it is driven by a fundamentally different objective. Wang et al. treat the variance they uncover as a systemic flaw, a group-level bias that should be measured and, ideally, corrected to ensure uniformity. Our work adopts their partitioned modeling approach but shifts the unit of analysis from the collective group (a region) to the most fundamental decision-maker: the individual judge. We therefore re-interpret the resulting variance not as a systemic flaw, but as a stable, predictable, and measurable feature of individual judicial behavior. Their work powerfully demonstrates that where a case is heard matters; our study takes the critical next step to prove that who hears it matters even more.

A final recent paper by Barale, Rovatsos and Bhuta [63] confirms this idea. The main focus of the authors is to challenge the capacity of machine learning to meaningfully evaluate legal fairness. They correctly diagnose that naive statistical models are blind to legal context and that the disparity they detect cannot be simply equated with unfairness. Disparity of outcomes does not automatically imply unfairness, if the disparity can be explained by a legitimate cause. They employ a multi-method approach. First, a feature-based statistical analysis reveals massive outcome disparities in asylum grant rate correlated with judge identity, date and court location.

Second, the authors use predictive modeling to confirm these results. While their models achieve high accuracy, they discover that this performance is not driven by legal reasoning. Their analysis explicitly shows that “Legal explanations, including credibility assessments and grounds for persecution, consistently rank lower in predictive importance” than institutional metadata. The models learned to replicate the system's patterns by relying on illegitimate, context-based signals like the date of the hearing and, most powerfully, the identity of the judge.

To isolate and magnify the effect of judicial behavior, the authors constructed a controlled subset. The authors chose decisions from six specific judges who represent the extreme ends of the decision-making spectrum: three with consistently high asylum grant rates (the "HGR" group) and three with consistently low, or even zero, grant rates (the "LGR" group). By removing the noise from judges with moderate or inconsistent records, this subset creates a simplified, laboratory-like environment to test how judicial patterns influence outcomes. This high-contrast dataset served two main, powerful purposes in their analysis.

The first goal was to see if certain types of claimants were being funneled to certain types of judges. The analysis of this subset yielded a critical finding: vulnerable groups were not randomly distributed. The authors state, “our study on the controlled subset shows these groups are disproportionately assigned to the LGR (Low Grant Rate) judges”. This reveals a potential systemic bias in the case allocation process itself. The devastating impact of this bias is quantified by their finding that when these same cases are heard by HGR judges instead, “outcomes improve substantially, +20% for minors, +54% for LGBTQIA+ claimants”. This demonstrates that the case outcome is less dependent on the merits of the claim and more on the arbitrary luck of which judge is assigned.

The second purpose was to test their machine learning model in an environment where judicial behavior was extremely predictable. If their theory was correct—that the model was primarily learning the judge's habits rather than the legal substance of a case—then its performance should improve on this simplified dataset. The results confirmed this hypothesis unequivocally. The full model's F1-score, a measure of accuracy, jumped from 89.8% on the complete dataset to 93.8% on the controlled subset. This increased accuracy doesn't mean the model became "smarter" about the law; it means that with the ambiguity of moderate judges removed, it became exceptionally good at learning the clear, binary patterns of the LGR and HGR judges. This confirms that the model's high performance is largely an illusion, driven by its ability to replicate institutional biases rather than engage in legal reasoning.

Ultimately, the study by Barale, Rovatsos and Bhuta provides compelling evidence for the existence of a judge variable. By demonstrating that legally salient factors are less predictive than extra-legal ones, they reveal a system where fairness is critically compromised. Since there is often no single true outcome to be found, their work suggests that the most rational way to predict a case's result is not to rely only on features representative of legal merits, but to include decision-making patterns of the judges holding the power to decide.

## 5. Experimental design

For our predictive models, we implemented three common machine learning algorithms in the SciKit Library : Random Forest (RF), Extreme Gradient Boost (XGB) and Support Vector Classification. These algorithms were selected for their simplicity, as the goal of this paper is not to propose new computation techniques, or achieve best performance in the field of Legal Judgment Prediction. The models were trained to predict three possible custody outcomes: primary maternal residence, primary paternal residence, or shared custody arrangements. By comparing the performance metrics of judge-specific models (the model is trained on a single judge) against the generic model (the model is trained on a pool of judges), we can quantify the extent to which profiling the judge will enhance predictions.

To approximate real-world usage as closely as possible, our model uses only case features that would realistically be accessible before a judicial decision is made. All variables are categorical, mostly binary, and deliberately neutral with respect to the decision outcome. This neutrality prevents the model from detecting implicit cues in the text structure or judgment phrasing itself. This design responds to concerns raised by Medvedeva and McBride [64], who argue for a critical distinction between outcome identification, outcome categorisation, and outcome forecasting in LJP. They argue that valid LJP systems must rely strictly on data available to users prior to judgment, avoiding reliance on post-decision information. Our approach builds on this foundation, grounded in the premise that similar legal inputs tend to produce consistent judicial outcomes all other contextual variables remaining the same (the ceteris paribus condition). By explicitly linking past judicial behavior to case features, we evaluate whether encoding individual decision patterns enhances the system's ability to anticipate outcomes under comparable conditions. This study investigates whether incorporating individual judicial decision-making patterns—derived from each judge's historical rulings—enhances the predictive accuracy of legal judgment models. Current approaches in LJP operate under the implicit assumption that judges are functionally interchangeable, disregarding judicial variability as a meaningful variable in shaping outcomes. By explicitly integrating judge-specific behavioral profiles into model training, we test whether personalized judicial profiles refine outcome forecasts. Our research challenges the prevailing abstraction of judges as uniform decision-makers and advances a more nuanced understanding of how judge-profiles interact with facts of the case to determine the outcome.

**In-domain testing**- We want to compare specialist models performance to generalist model performance, because we hypothesized that a generalist model would struggle to capture a generic pattern as if judges were perfect substitutes one for another. In-domain tests validate whether specialized models outperform the generalist in their "home" context. Each model, whether judge-specific (specialist) or generalist, is first evaluated on its own held-out test set, representing the same judicial context it was trained on. For instance: A specialist model trained on Judge A's historical rulings is tested on Judge A's reserved test cases. The generalist model (judge-agnostic), trained on aggregated data from the Generic Bucket (judges with <300 decisions), is tested on the same Generic Bucket's test set. This test answers: **How well does a model perform compared to the real-world judge it is trying to emulate?** The In-Domain validity test establishes whether models can reliably capture the ruling-profile of a specific judge. High performance here suggests the specialist model has successfully learned consistent patterns within a single judge ruling history.

**Cross-Domain generalization**- Cross-domain tests reveal if specialist models systematically fail on other judges' data, confirming judicial variability. Each specialist model is then evaluated on all other buckets' test sets, creating a dense performance matrix. The generalist model is also tested on every judge-specific test set. This test answers: **Are decision-making patterns learned from one judge transferable to others?** It exposes whether models trained on specific judges exhibit portable logic that aligns with broader judicial tendencies (e.g., does Judge A's model generalize well to Judge B's cases?). Crucially, this setup creates a "confusion matrix" of models vs. judges, where performance discrepancies highlight judicial uniqueness: If a specialist model fails on other judges' data, it signals that judge's decision-making is distinct. Conversely, if models (including the generalist) perform consistently across multiple judges, it suggests that shared patterns carry moreweight than idiosyncratic tendencies.

At the data ingestion stage, case features, judge features and outcomes of the cases are merged and one-hot encoded to categorical variables. To capture variation in judicial behaviour, the dataset is partitioned into judge-specific buckets for magistrates with sufficient case volume and a residual "Generic" bucket that aggregates all remaining decisions. The "Generic Bucket" refers to the aggregation of cases assigned to judges with fewer than 300 decisions, serving as a pooled baseline for generalist model performance.

Class imbalance is addressed through configurable random oversampling and undersampling, applied only to the training data to prevent contamination of validation and test sets. To address class imbalance during training, we applied random undersampling and oversampling to the combined training and validation sets. Specifically, the majority class "mother" (class 0) was undersampled to 34% (run1) or 40% (run2) of the training data, while the two minority classes "father" (class 1) and "shared" custody (class 2) were each oversampled to 33% (run1) or 30% (run2). This balancing strategy ensures that the model is exposed to a more uniform class distribution during learning, improving its ability to generalise across all outcome types.

We performed hyperparameter tuning using stratified 5-fold cross-validation. A strict evaluation protocol is followed: 20% of each bucket is withheld as an untouched test set to provide unbiased performance estimates. In evaluating model performance, we adopt standard classification metrics (accuracy, precision, recall, and F1-score) per class and macro. Given the pronounced class imbalance in our dataset, we used macro-averaged metrics. All reported evaluation metrics are macro-averaged and expressed as percentages, ensuring equal weight is given to each class regardless of frequency. As noted in a 2022 LJP survey by Feng, Li and Ng [65], micro-averaged metrics can be misleading in imbalanced settings, as they disproportionately reflect the performance on majority classes. Macro F1 computes the F1-score separately for each class and averages the results without weighting by class frequency, ensuring that each outcome contributes equally to the final score. As argued by Lage-Freitas [66], macro metrics are more appropriate in legal judgment prediction tasks, as they reward models that genuinely learn to discriminate between all classes rather than defaulting to dominant patterns. In addition, feature importance in the Random Forest is computed via impurity reduction measured using the Gini index, averaged across all decision trees.

# 6. In-Domain Validity

### a *Random Forest*

**Table 11. Random Forest (Run1).**
Hyperparameters: Estimators [100-300], Max depth 7, Bootstrap: true.

| Bucket | n_test | Distr. % | Accur % | Precis % | Recall % | F1-CV % | std % | F1-test % |
|---|---|---|---|---|---|---|---|---|
| Hauynes | 100 | 64/25/11 | 79 | 73.06 | 86.55 | 75 | 3.4 | 76.19 |
| Cabeche | 147 | 63/21/14 | 79.59 | 74.3 | 84.62 | 79.71 | 1.83 | 77.14 |
| Babiche | 154 | 66/23/10 | 83.11 | 74.76 | 83.87 | 75.08 | 4.04 | 78.04 |
| Anatole | 165 | 67/23/09 | 83.03 | 75.36 | 85.82 | 78.05 | 2.82 | 78.61 |
| Generic | 2376 | 68/21/10 | 83.79 | 74.88 | 85.59 | 78.92 | 0.79 | 78.7 |
| Dacrons | 127 | 64/24/11 | 83.46 | 75.99 | 85.5 | 84.19 | 2.64 | 79.22 |
| Jobelin | 82 | 63/24/12 | 85.36 | 78.63 | 83.84 | 80.59 | 7.74 | 80.55 |
| Faubers | 110 | 63/27/09 | 81.81 | 76.68 | 87.93 | 75 | 3.4 | 80.66 |
| Labourg | 62 | 74/14/11 | 87.09 | 78.41 | 87.18 | 83.35 | 4.91 | 81.9 |
| Echevin | 118 | 61/26/11 | 86.44 | 80.57 | 87.6 | 86.4 | 2.44 | 83.2 |
| Inconel | 82 | 62/25/12 | 86.58 | 82.18 | 88.26 | 84.64 | 6.49 | 84.67 |
| Gargote | 115 | 74/18/06 | 90.43 | 81.23 | 91.95 | 78.21 | 6.29 | 85.58 |
| Kochias | 72 | 58/25/16 | 87.5 | 84.25 | 89.81 | 83.1 | 4.57 | 85.84 |
| Orillon | 63 | 65/23/11 | 90.47 | 84.06 | 91.17 | 88.93 | 5.29 | 87.01 |

**Table 12. Random Forest (Run2).**
Hyperparameters: Estimators [100-300], Max depth 7, Bootstrap: true.

| Bucket | n_test | Distr. % | Accur % | Precis % | Recall % | F1-CV % | std % | F1-test % |
|---|---|---|---|---|---|---|---|---|
| Jobelin | 82 | 63/24/12 | 85.36 | 78.63 | 83.84 | 81.63 | 7.39 | 80.55 |
| Labourg | 62 | 74/14/11 | 87.09 | 78.41 | 87.18 | 86.69 | 3.33 | 81.9 |
| Generic | 2376 | 68/21/10 | 87.45 | 79.65 | 85.74 | 82.29 | 1.49 | 82.28 |
| Babiche | 154 | 66/23/10 | 87.01 | 80.25 | 85.83 | 79.16 | 3.04 | 82.59 |
| Dacrons | 127 | 64/24/11 | 87.4 | 79.88 | 88.2 | 86.21 | 4.13 | 82.73 |
| Faubers | 110 | 63/27/09 | 84.54 | 79.87 | 88.09 | 78.15 | 2.02 | 83.14 |
| Anatole | 165 | 67/23/09 | 87.87 | 81.02 | 87.11 | 81.63 | 2.88 | 83.5 |
| Kochias | 72 | 58/25/16 | 87.5 | 84.16 | 85.84 | 87.21 | 4.45 | 84.68 |
| Inconel | 82 | 62/25/12 | 86.58 | 83.47 | 87.32 | 84.07 | 6.37 | 85.17 |
| Cabeche | 147 | 63/21/14 | 88.43 | 83.18 | 89.27 | 82.03 | 2.26 | 85.63 |
| Gargote | 115 | 74/18/06 | 91.3 | 82.5 | 91.14 | 82.49 | 7.51 | 86.23 |
| Hauynes | 100 | 64/25/11 | 89 | 82.43 | 94.27 | 76.97 | 3.54 | 86.35 |
| Echevin | 118 | 61/26/11 | 88.98 | 86.25 | 87.74 | 88.8 | 1.99 | 86.96 |
| Orillon | 63 | 65/23/11 | 93.65 | 88.03 | 92.79 | 89.35 | 4.56 | 90 |

**Table 13. Random Forest Confusion Matrix for run 1 and run 2.**

| | | Worst Model: Hauynes | | | | Generic Model | | | | Best Model: Orillon | | | |
|---|---|---|---|---|---|---|---|---|---|---|---|---|---|
| Run# | Class | Pred 0 | Pred 1 | Pred 2 | total | Pred 0 | Pred 1 | Pred 2 | total | Pred 0 | Pred 1 | Pred 2 | total |
| 1 | True 0 | 44 | 10 | 10 | 64 | 1307 | 162 | 156 | 1625 | 36 | 2 | 3 | 41 |
| 1 | True 1 | 0 | 25 | 0 | 25 | 18 | 493 | 2 | 513 | 0 | 15 | 0 | 15 |
| 1 | True 2 | 1 | 0 | 10 | 11 | 33 | 14 | 191 | 238 | 1 | 0 | 6 | 7 |
| 1 | total | 45 | 35 | 20 | 100 | 1358 | 669 | 349 | 2376 | 37 | 17 | 9 | 63 |

| | | Worst Model: Jobelin | | | | Generic Model | | | | Best Model: Orillon | | | |
|---|---|---|---|---|---|---|---|---|---|---|---|---|---|
| Run# | Class | Pred 0 | Pred 1 | Pred 2 | total | Pred 0 | Pred 1 | Pred 2 | total | Pred 0 | Pred 1 | Pred 2 | total |
| 2 | True 0 | 45 | 2 | 5 | 52 | 1419 | 103 | 103 | 1625 | 38 | 0 | 3 | 41 |
| 2 | True 1 | 2 | 17 | 1 | 20 | 36 | 475 | 2 | 513 | 0 | 15 | 0 | 15 |
| 2 | True 2 | 1 | 1 | 8 | 10 | 42 | 12 | 184 | 238 | 1 | 0 | 6 | 7 |
| 2 | total | 48 | 20 | 14 | 72 | 1497 | 590 | 289 | 2376 | 39 | 15 | 9 | 63 |

### *b Extreme Gradient Boost (XGB)*

**Table 14. XGB Models (Run 1).**

HyperParameters : Estimators [100-250], Max depth 7. Min. Child Weight 5, Learning rate [0.05 – 0.2], Gamma: [0.85 – 1.0], Sub Sample: [0.5 – 0.8], Features Sample [0.5 – 0.8].

| Bucket | n_test | Distr. % | Accur % | Precis % | Recall % | F1-CV % | std % | F1-test % |
|---|---|---|---|---|---|---|---|---|
| **Hauynes** | 100 | 64/25/11 | 76.00 | 69.26 | 80.04 | 73.95 | 3.22 | **72.09** |
| **Cabeche** | 147 | 63/21/14 | 80.27 | 74.16 | 82.66 | 79.66 | 2.33 | **76.92** |
| **Dacrons** | 157 | 64/24/11 | 83.46 | 74.97 | 82.22 | 82.05 | 4.39 | **77.65** |
| **Jobelin** | 82 | 63/24/12 | 84.14 | 76.89 | 80.51 | 75.99 | 4.83 | **78.32** |
| **Faubers** | 110 | 63/27/09 | 80.90 | 74.67 | 87.46 | 75.06 | 3.68 | **78.77** |
| **Generic** | **2376** | **68/21/10** | **85.10** | **76.16** | **87.07** | **80.95** | **0.50** | **80.00** |
| **Labourg** | 62 | 74/14/11 | 85.48 | 75.66 | 89.44 | 80.56 | 2.71 | **80.59** |
| **Echevin** | 118 | 61/26/11 | 84.74 | 78.81 | 86.69 | 85.97 | 2.69 | **81.23** |
| **Anatole** | 165 | 67/23/09 | 86.06 | 78.47 | 85.66 | 77.82 | 3.14 | **81.46** |
| **Orillon** | 63 | 65/23/11 | 87.30 | 80.39 | 84.18 | 85.88 | 7.46 | **81.94** |
| **Inconel** | 82 | 62/25/12 | 84.14 | 79.21 | 86.95 | 82.87 | 6.01 | **82.11** |
| **Babiche** | 154 | 66/23/10 | 87.01 | 80.97 | 85.83 | 76.54 | 3.31 | **82.94** |
| **Kochias** | 72 | 58/25/16 | 84.72 | 82.85 | 85.18 | 83.38 | 4.92 | **83.16** |
| **Gargote** | 115 | 74/18/06 | 91.30 | 84.06 | 91.14 | 80.66 | 5.31 | **87.14** |

**Table 15. XGB Models (Run 2).**

HyperParameters : Estimators [120-250], Max depth 7. Min. Child Weight 5, Learning rate [0.05 – 0.2], Gamma: [0.85 – 1.0], Sub Sample: [0.5 – 0.8], Features Sample [0.5 – 0.8].

| Bucket | n_test | Distr. % | Accur % | Precis % | Recall % | F1-CV % | std % | F1-test % |
|---|---|---|---|---|---|---|---|---|
| **Faubers** | 110 | 63/27/09 | 76.36 | 69.93 | 79.68 | 75.06 | 3.68 | **73.32** |
| **Hauynes** | 100 | 64/25/11 | 79.00 | 71.95 | 81.60 | 73.95 | 3.22 | **74.61** |
| **Jobelin** | 82 | 63/24/12 | 84.14 | 76.89 | 80.51 | 75.99 | 4.83 | **78.32** |
| **Dacrons** | 157 | 64/24/11 | 85.03 | 74.67 | 83.04 | 82.05 | 4.39 | **78.93** |
| **Cabeche** | 147 | 63/21/14 | 84.35 | 79.05 | 83.92 | 79.66 | 2.33 | **81.11** |
| **Anatole** | 165 | 67/23/09 | 87.27 | 79.80 | 85.15 | 77.82 | 3.14 | **82.10** |
| **Inconel** | 82 | 62/25/12 | 84.14 | 79.21 | 86.95 | 82.87 | 6.01 | **82.11** |
| **Generic** | **2376** | **68/21/10** | **87.50** | **79.23** | **87.83** | **80.95** | **0.50** | **82.63** |
| **Labourg** | 62 | 74/14/11 | 88.70 | 79.74 | 90.89 | 80.56 | 2.71 | **84.22** |
| **Babiche** | 154 | 66/23/10 | 88.96 | 82.88 | 86.81 | 76.54 | 3.31 | **84.61** |
| **Echevin** | 118 | 61/26/11 | 88.13 | 83.18 | 87.90 | 85.97 | 2.69 | **85.23** |
| **Kochias** | 72 | 58/25/16 | 87.50 | 84.93 | 87.83 | 83.38 | 4.92 | **85.82** |
| **Gargote** | 115 | 74/18/06 | 92.17 | 83.62 | 91.53 | 80.66 | 5.31 | **87.09** |
| **Orillon** | 63 | 65/23/11 | 92.06 | 88.09 | 86.62 | 85.88 | 7.46 | **87.31** |

**Table 16. XGBoost Confusion Matrix for run 1 and run 2.**

| | | Worst Model: Hauynes | | | | Generic Model | | | | Best Model: Gargote | | | |
|---|---|---|---|---|---|---|---|---|---|---|---|---|---|
| Run# | Class | Pred 0 | Pred 1 | Pred 2 | total | Pred 0 | Pred 1 | Pred 2 | total | Pred 0 | Pred 1 | Pred 2 | total |
| 1 | True 0 | 45 | 8 | 11 | 64 | 1336 | 123 | 166 | 1625 | 78 | 6 | 2 | 86 |
| 1 | True 1 | 3 | 22 | 0 | 25 | 19 | 485 | 9 | 513 | 1 | 20 | 0 | 21 |
| 1 | True 2 | 1 | 1 | 9 | 11 | 25 | 12 | 201 | 238 | 1 | 0 | 7 | 8 |
| 1 | total | 49 | 31 | 20 | 100 | 1380 | 620 | 376 | 2376 | 80 | 26 | 9 | 115 |

| | | Worst Model: Faubers | | | | Generic Model | | | | Best Model: Orillon | | | |
|---|---|---|---|---|---|---|---|---|---|---|---|---|---|
| Run# | Class | Pred 0 | Pred 1 | Pred 2 | total | Pred 0 | Pred 1 | Pred 2 | total | Pred 0 | Pred 1 | Pred 2 | total |
| 2 | True 0 | 53 | 10 | 7 | 70 | 1402 | 99 | 124 | 1625 | 39 | 0 | 2 | 41 |
| 2 | True 1 | 8 | 22 | 0 | 30 | 28 | 476 | 9 | 513 | 1 | 14 | 0 | 15 |
| 2 | True 2 | 0 | 1 | 9 | 10 | 26 | 11 | 201 | 238 | 2 | 0 | 5 | 7 |
| 2 | total | 61 | 33 | 16 | 110 | 1456 | 586 | 334 | 2376 | 43 | 14 | 7 | 63 |

### c Support Vector Classification

**Table 17. Support Vector Classification (Run 1).**

HyperParameters : Degree [2-4]. Kernel [poly, linear, rbf], Gamma [scale, auto], Coef0 [0.05, 0.1, 0.5], Class Weight balanced, C [1, 10, 100].

| Bucket | n_test | Distr. % | Accur % | Precis % | Recall % | F1-CV % | std % | F1-test % |
|---|---|---|---|---|---|---|---|---|
| **Dacrons** | 157 | 64/24/11 | 83.46 | 75.34 | 79.58 | 82.63 | 2.13 | 77.16 |
| **Inconel** | 82 | 62/25/12 | 80.48 | 75.10 | 83.12 | 85.13 | 5.20 | 78.00 |
| **Cabeche** | 147 | 63/21/14 | 81.63 | 75.76 | 84.53 | 79.99 | 4.87 | 78.63 |
| **Generic** | **2376** | **68/21/10** | **85.69** | **77.19** | **86.02** | **80.57** | **1.07** | **80.35** |
| **Anatole** | 165 | 67/23/09 | 87.87 | 81.94 | 84.64 | 83.64 | 2.77 | 83.06 |
| **Faubers** | 110 | 63/27/09 | 85.45 | 79.66 | 89.84 | 72.74 | 2.87 | 83.47 |
| **Kochias** | 72 | 58/25/16 | 86.11 | 83.71 | 85.97 | 84.24 | 4.96 | 83.99 |
| **Echevin** | 118 | 61/26/11 | 86.44 | 84.01 | 87.05 | 87.65 | 2.16 | 85.40 |
| **Gargote** | 115 | 74/18/06 | 91.30 | 82.50 | 91.14 | 81.11 | 6.70 | 86.23 |
| **Babiche** | 154 | 66/23/10 | 90.25 | 86.05 | 87.47 | 80.19 | 4.87 | 86.54 |
| **Hauynes** | 100 | 64/25/11 | 89.00 | 82.67 | 94.27 | 74.74 | 3.81 | 86.69 |
| **Jobelin** | 82 | 63/24/12 | 90.24 | 84.48 | 91.15 | 78.28 | 5.78 | 87.22 |
| **Labourg** | 62 | 74/14/11 | 91.93 | 84.83 | 92.33 | 81.08 | 4.14 | 88.12 |
| **Orillon** | 63 | 65/23/11 | 95.23 | 90.00 | 97.56 | 86.31 | 3.34 | 92.85 |

**Table 18. Support Vector Classification (Run 2).**

HyperParameters : Degree [2-4]. Kernel [poly, linear, rbf], Gamma [scale, auto], Coef0 [0.05, 0.1, 0.5], Class Weight balanced, C [1, 10, 100].

| Bucket | n_test | Distr. % | Accur % | Precis % | Recall % | F1-CV % | std % | F1-test % |
|---|---|---|---|---|---|---|---|---|
| **Dacrons** | 157 | 64/24/11 | 84.25 | 76.61 | 81.96 | 83.56 | 2.34 | 78.85 |
| **Hauynes** | 100 | 64/25/11 | 81.00 | 75.47 | 90.10 | 77.25 | 3.86 | 78.95 |
| **Inconel** | 82 | 62/25/12 | 82.92 | 78.41 | 80.82 | 84.83 | 4.44 | 79.53 |
| **Faubers** | 110 | 63/27/09 | 83.63 | 77.67 | 85.39 | 75.70 | 3.17 | 80.73 |
| **Cabeche** | 147 | 63/21/14 | 83.67 | 77.97 | 85.40 | 81.09 | 5.88 | 80.76 |
| **Generic** | **2376** | **68/21/10** | **86.61** | **78.45** | **85.74** | **81.67** | **0.65** | **81.38** |
| **Kochias** | 72 | 58/25/16 | 84.72 | 82.08 | 83.20 | 86.01 | 4.82 | 82.06 |
| **Anatole** | 165 | 67/23/09 | 87.87 | 82.09 | 84.08 | 82.93 | 3.28 | 82.97 |
| **Jobelin** | 82 | 63/24/12 | 89.02 | 83.45 | 85.12 | 82.71 | 7.37 | 84.21 |
| **Echevin** | 118 | 61/26/11 | 86.44 | 83.78 | 87.67 | 87.03 | 2.20 | 85.54 |
| **Babiche** | 154 | 66/23/10 | 90.25 | 86.05 | 87.47 | 80.34 | 5.20 | 86.54 |
| **Orillon** | 63 | 65/23/11 | 90.47 | 84.95 | 89.76 | 86.63 | 4.13 | 86.54 |
| **Gargote** | 115 | 74/18/06 | 93.04 | 84.83 | 91.92 | 81.50 | 8.93 | 87.97 |
| **Labourg** | 62 | 74/14/11 | 91.93 | 84.83 | 92.33 | 83.42 | 3.45 | 88.12 |

**Table 19. Support Vector Classification Confusion Matrix for run 1 and run 2.**

| | | Worst Model: Dacrons | | | | Generic Model | | | | Best Model: Orillon | | | |
|---|---|---|---|---|---|---|---|---|---|---|---|---|---|
| Run# | Class | Pred 0 | Pred 1 | Pred 2 | total | Pred 0 | Pred 1 | Pred 2 | total | Pred 0 | Pred 1 | Pred 2 | total |
| 1 | True 0 | 69 | 6 | 7 | 82 | 1380 | 88 | 157 | 1625 | 38 | 0 | 3 | 41 |
| 1 | True 1 | 3 | 28 | 0 | 31 | 43 | 455 | 15 | 513 | 0 | 15 | 0 | 15 |
| 1 | True 2 | 3 | 2 | 9 | 14 | 30 | 7 | 201 | 238 | 0 | 0 | 7 | 7 |
| 1 | total | 75 | 36 | 16 | 127 | 1453 | 550 | 373 | 2376 | 38 | 15 | 10 | 63 |
| | | Worst Model: Dacrons | | | | Generic Model | | | | Best Model: Labourg | | | |
| Run# | Class | Pred 0 | Pred 1 | Pred 2 | total | Pred 0 | Pred 1 | Pred 2 | total | Pred 0 | Pred 1 | Pred 2 | total |
| 2 | True 0 | 69 | 6 | 7 | 82 | 1410 | 98 | 117 | 1625 | 42 | 2 | 2 | 46 |
| 2 | True 1 | 3 | 28 | 0 | 31 | 45 | 452 | 16 | 513 | 0 | 9 | 0 | 9 |
| 2 | True 2 | 2 | 2 | 10 | 14 | 35 | 7 | 196 | 238 | 1 | 0 | 6 | 7 |
| 2 | total | 74 | 36 | 17 | 127 | 1490 | 557 | 329 | 2376 | 43 | 11 | 8 | 62 |

### d Discussion

In-domain validity tests results provide compelling evidence in support of the central hypothesis: judge identity matters in predicting legal case outcomes. For both runs, the generalist models performed robustly, scoring an F1 of 78.70% - 82.28% (RF), 80.00% - 82.63% (XGB) and 80.35% - 81.38% (SVC). Apparently, this strong baseline challenges the hypothesis suggesting that judge-agnostic models can still perform competitively. While it is worth noting that not all specialist models outperform the generalist, several specialists models outperformed the generalist models despite the strong baseline. The trend emerges clearly: best models are specialist. Most specialist models are consistently stronger than the generalist model, suggesting that each model has successfully captured stable individual patterns. The fact that these patterns emerge across Random Forest, XGBoost, and SVC points to the robustness of the underlying signal in the data. The fact that different algorithms converge on similar conclusions lends strong empirical weight to the legal realist hypothesis. Both runs confirm the stability of the modeling approach under different class distributions. The relatively low standard deviations across cross-validation folds for most judges further support the internal consistency of these patterns. Observing the confusion matrices, we can evaluate not just overall effectiveness but also the nature of classification errors and the consistency across classes. The best-case specialists results are particularly impressive given the small sample size. For a relatively imbalanced three-class classification task, this level of precision across classes suggests that the model is not just memorizing majority class behavior (since the training data was balanced) but has internalized a meaningful representation of how this judge rules.

The experiment solidifies the empirical case for the importance of judicial identity in legal outcome prediction. While a generalist model performs well and captures broad trends, it cannot fully replicate the precision gained from modeling individual judicial behavior. These findings offer empirical support for legal realism. Legal outcome prediction is not simply a question of learning the law—it is also a matter of learning the judge. However, the performance of weaker specialist models also signals a caution: without sufficient data or stable patterns, judicial identity alone does not guarantee better prediction. In short, judicial identity is a powerful variable, but not an universal solution. Modeling the judge is not a magic bullet with poor quality data that do not represent consistent temporal series.

## 7. Cross-Domain Generalization

**Supporting the Realist Hypothesis**: Values in Dark Blue are specialist models that overperformed compared to generalist model. Values in Red exhibit a noticeable drop in F1 (strong signal : variation over 5%). Values in Purple exhibit a small drop in F1 (weak signal : variation less than 5%).

**Supporting the Legalist Hypothesis**: Values in Yellow are specialist models that underperformed compared to generalist model. Values in Green exhibit equal performance or improvement.

What matters here is not just the absolute score but the relative positioning. Reading the table horizontally, we can compare the In-Domain F1 value with Cross-Domain F1 values. According to the Realist Hypothesis, we would expect a sharp drop in performance out of domain. Reading the table vertically, we can compare the specialist model to the generalist model. According to the Realist Hypothesis, we would expect the F1 score for the specialist model to be higher than the generalist F1.

**Table 20. Random Forest Models Cross Comparison (Macro-F1).**

| Model | On A | On B | On C | On D | On E | On F | On G | On H | On I | On J | On K | On L | On O | on Gen |
|---|---|---|---|---|---|---|---|---|---|---|---|---|---|---|
| A >> | 83.50 | *76.28* | *69.71* | *73.14* | *80.09* | *69.47* | *66.90* | *78.39* | *78.87* | *84.79* | *76.06* | *80.89* | *80.29* | *77.24* |
| B >> | *75.88* | 82.59 | *73.90* | *78.79* | *83.56* | *68.93* | *76.63* | *79.10* | *75.07* | *75.58* | *75.60* | *78.34* | *77.50* | *77.32* |
| C >> | *74.20* | *76.69* | 85.63 | *82.03* | *79.65* | *72.51* | *72.29* | *72.27* | *78.47* | *85.33* | *78.90* | *78.38* | *81.78* | *77.61* |
| D >> | *82.34* | *78.56* | *79.40* | 82.73 | *86.59* | *71.30* | *87.15* | *81.27* | *83.15* | *78.38* | *83.12* | *81.96* | *88.45* | *82.22* |
| E >> | *86.68* | *75.93* | *79.93* | *83.07* | 86.96 | *71.95* | *81.76* | *83.25* | *82.39* | *83.78* | *85.38* | *78.38* | *86.83* | *79.48* |
| F >> | *75.38* | *75.63* | *76.08* | *74.54* | *71.96* | 83.14 | *79.27* | *76.73* | *80.56* | *74.49* | *92.14* | *76.34* | *80.97* | *76.43* |
| G >> | *75.26* | *72.31* | *73.08* | *82.51* | *77.30* | *80.58* | 86.23 | *80.36* | *80.39* | *83.51* | *84.27* | *67.97* | *80.21* | *79.52* |
| H >> | *77.72* | *78.23* | *75.05* | *78.12* | *80.43* | *76.73* | *78.83* | 86.35 | *85.24* | *89.21* | *84.07* | *71.62* | *78.15* | *77.81* |
| I >> | *74.40* | *72.34* | *76.73* | *78.02* | *80.40* | *68.50* | *66.37* | *80.55* | 85.17 | *78.65* | *81.03* | *72.93* | *83.85* | *76.61* |
| J >> | *72.94* | *74.20* | *74.94* | *80.86* | *71.48* | *75.38* | *82.40* | *76.97* | *75.86* | 80.55 | *78.48* | *65.49* | *73.05* | *77.59* |
| K >> | *79.58* | *78.26* | *80.53* | *84.58* | *85.97* | *71.02* | *76.12* | *81.27* | *82.21* | *78.15* | 84.68 | *76.63* | *86.83* | *81.79* |
| L >> | *70.92* | *78.93* | *81.45* | *83.36* | *80.57* | *75.43* | *77.36* | *81.85* | *77.60* | *75.79* | *85.52* | 81.90 | *81.15* | *77.15* |
| O >> | *72.95* | *75.05* | *85.21* | *84.70* | *84.16* | *80.31* | *83.14* | *83.19* | *78.38* | *79.69* | *79.26* | *82.08* | 90.00 | *79.82* |
| Gen. | *81.14* | *80.17* | *75.75* | *79.60* | *80.49* | *68.59* | *76.88* | *81.86* | *79.46* | *79.41* | *84.48* | *80.28* | *81.78* | 82.28 |

**Table 21. XGBoost Models Cross Comparison (Macro-F1).**

| Model | On A | On B | On C | On D | On E | On F | On G | On H | On I | On J | On K | On L | On O | on Gen |
|---|---|---|---|---|---|---|---|---|---|---|---|---|---|---|
| A >> | 82.21 | 76.43 | 71.05 | 78.50 | 79.89 | 68.05 | 72.02 | 77.29 | 77.06 | 75.68 | 75.85 | 80.79 | 83.85 | 75.36 |
| B >> | 71.87 | 84.61 | 76.26 | 77.32 | 81.44 | 68.95 | 75.91 | 74.14 | 74.73 | 72.65 | 75.43 | 77.08 | 79.91 | 76.65 |
| C >> | 73.80 | 77.14 | 81.11 | 80.91 | 77.06 | 73.43 | 75.58 | 72.78 | 77.46 | 73.12 | 84.35 | 80.28 | 79.62 | 78.40 |
| D >> | 83.50 | 79.11 | 78.06 | 78.93 | 83.56 | 68.79 | 82.33 | 83.24 | 79.62 | 81.78 | 81.82 | 84.04 | 82.63 | 81.45 |
| E >> | 83.75 | 79.70 | 79.34 | 82.13 | 85.23 | 74.44 | 78.17 | 79.03 | 81.06 | 76.33 | 81.61 | 76.63 | 83.47 | 81.75 |
| F >> | 73.50 | 75.24 | 73.69 | 77.44 | 77.87 | 73.32 | 73.54 | 74.14 | 76.37 | 79.62 | 85.47 | 74.86 | 77.47 | 75.58 |
| G >> | 77.59 | 77.73 | 75.02 | 83.07 | 74.82 | 77.45 | 87.09 | 76.77 | 76.86 | 84.57 | 88.01 | 70.07 | 78.71 | 79.86 |
| H >> | 74.76 | 76.06 | 70.19 | 75.30 | 75.41 | 72.57 | 78.05 | 74.61 | 78.73 | 86.32 | 83.64 | 79.47 | 76.74 | 77.79 |
| I >> | 74.58 | 76.37 | 78.18 | 79.65 | 77.82 | 70.04 | 71.35 | 76.19 | 82.11 | 77.64 | 84.36 | 78.86 | 83.85 | 76.79 |
| J >> | 68.11 | 73.99 | 71.87 | 74.17 | 69.24 | 68.36 | 70.45 | 74.25 | 71.59 | 78.32 | 73.97 | 75.23 | 73.05 | 74.95 |
| K >> | 76.68 | 78.23 | 78.71 | 82.21 | 81.43 | 75.41 | 75.87 | 80.88 | 72.33 | 86.83 | 85.82 | 73.52 | 80.21 | 79.22 |
| L >> | 70.92 | 78.93 | 81.45 | 83.36 | 80.57 | 75.43 | 77.36 | 81.85 | 77.60 | 75.79 | 85.52 | 81.90 | 81.15 | 77.15 |
| O >> | 72.91 | 73.53 | 82.32 | 81.47 | 80.23 | 73.41 | 82.49 | 78.26 | 76.80 | 79.60 | 77.19 | 76.10 | 87.31 | 78.82 |
| Gen. | 77.51 | 78.39 | 75.92 | 80.68 | 84.74 | 75.88 | 78.69 | 80.11 | 71.19 | 83.15 | 83.20 | 77.20 | 78.15 | 82.63 |

**Table 22. SVC Models Cross Comparison (Macro-F1).**

| Model | On A | On B | On C | On D | On E | On F | On G | On H | On I | On J | On K | On L | On O | on Gen |
|---|---|---|---|---|---|---|---|---|---|---|---|---|---|---|
| A >> | 82.97 | 69.17 | 65.76 | 71.29 | 70.14 | 65.98 | 71.55 | 76.16 | 80.55 | 73.52 | 81.53 | 75.39 | 84.69 | 74.35 |
| B >> | 71.55 | 86.54 | 74.04 | 76.89 | 76.14 | 67.99 | 75.91 | 73.93 | 71.50 | 68.49 | 76.75 | 76.15 | 76.77 | 74.16 |
| C >> | 71.74 | 70.49 | 80.76 | 81.06 | 74.32 | 67.61 | 68.71 | 69.48 | 72.35 | 78.01 | 78.90 | 75.03 | 76.63 | 75.24 |
| D >> | 74.72 | 77.68 | 76.79 | 78.85 | 80.18 | 67.19 | 84.31 | 73.81 | 76.05 | 81.36 | 78.91 | 80.59 | 81.70 | 76.93 |
| E >> | 80.87 | 75.61 | 74.14 | 83.03 | 85.54 | 77.82 | 79.50 | 75.89 | 76.92 | 77.84 | 84.96 | 77.20 | 85.00 | 78.04 |
| F >> | 65.04 | 71.71 | 69.64 | 67.54 | 67.30 | 80.73 | 66.44 | 74.57 | 69.62 | 70.57 | 79.92 | 70.35 | 71.90 | 71.42 |
| G >> | 72.33 | 72.77 | 70.29 | 78.80 | 74.48 | 74.32 | 87.97 | 77.78 | 76.94 | 81.16 | 86.00 | 72.08 | 77.88 | 77.89 |
| H >> | 72.31 | 71.23 | 67.13 | 73.78 | 74.71 | 73.59 | 74.53 | 78.95 | 77.70 | 72.00 | 84.11 | 71.53 | 78.21 | 73.22 |
| I >> | 70.81 | 74.31 | 78.78 | 76.32 | 76.54 | 69.04 | 69.35 | 72.00 | 79.53 | 86.45 | 82.81 | 73.04 | 83.85 | 73.86 |
| J >> | 60.67 | 68.96 | 64.70 | 74.43 | 72.14 | 67.42 | 59.59 | 69.67 | 64.24 | 84.21 | 64.92 | 74.46 | 71.54 | 69.38 |
| K >> | 70.00 | 74.58 | 77.81 | 82.30 | 81.41 | 75.66 | 70.58 | 77.96 | 71.91 | 75.88 | 82.06 | 74.64 | 74.98 | 76.01 |
| L >> | 64.85 | 77.24 | 77.66 | 82.67 | 79.94 | 67.31 | 75.32 | 74.19 | 68.20 | 63.79 | 78.70 | 88.12 | 85.43 | 73.70 |
| O >> | 64.23 | 70.54 | 67.24 | 78.75 | 77.61 | 67.88 | 72.13 | 75.48 | 72.61 | 65.80 | 73.66 | 71.50 | 86.54 | 69.22 |
| Gen. | 73.76 | 80.76 | 74.77 | 80.96 | 79.00 | 65.83 | 79.63 | 81.37 | 72.43 | 66.98 | 82.30 | 80.79 | 88.59 | 81.38 |

While not uniform, the performance matrix reveals a consistent pattern. The cross-domain results offer strong empirical support for the hypothesis that judge's identity significantly influences legal case outcomes. The specialist models performed better than the generalist model on their own test sets: 13 times out of 13 for the Random Forest models, 9 times out of 13 for the XGB models, and 9 times out of 13 for the SVC models. Conversely, specialist models experienced drops in F1 performance (strong and weak signal) when applied to test sets ruled by other judges and on the generic test. There are some anomalies with dysfunctional XGB specialist models (Dacrons, Faubers and Hauynes), and dysfunctional SVC model (Dacrons). The generalist model remains more stable than specialist models across all test sets. It performs reliably on most judges, achieving accuracy scores comparable to many specialist models. The generalist model delivers steady and competitive performance across all test sets. However, while its average performance is solid, it never reaches the peak accuracies of the best specialist models. This reinforces the idea that generalist models are useful for broad prediction tasks, but specialist models are better suited for capturing the nuances of individual judicial behavior. Sharp declines across domains demonstrate that judges follow idiosyncratic rules that are sufficiently individualized to resist substitution. It is an empirical demonstration of judicial distinctiveness. Overall, these empirical results confort the Realist Hypothesis.

## 8. Top feature importance

We present a selection of Top 15 significant features across Random Forest Models. For better readability, features are collapsed in three clusters : Cluster 1 - Parents' Petitions and Child Preferences (features 12, 13, 14 and 16), Cluster 2 - Behavioral and Parenting Issues (features 18 – 31) and Cluster 3 - Practical Considerations (features 32 – 57).

In Scikit-learn's RandomForestClassifier and RandomForestRegressor, feature importance is computed by aggregating the decrease in node impurity—such as Gini impurity or entropy—attributable to each feature across all decision trees in the ensemble. At each split, the algorithm quantifies the reduction in uncertainty achieved by using a given feature, and these reductions are summed across the forest to reflect the feature's overall contribution to predictive performance. The resulting importance scores are then normalized so that the total importance across all features equals one. We only display the weights of the Top 15 features, which explains why they do not add up to 100%.

**Table 23. Top 15 Features Importance (Random Forest Model)**

| | Cluster 1 Petitions (%) | Cluster 2 Behavior (%) | Cluster 3 Practical (%) | Total (%) |
|---|---|---|---|---|
| Anatole | 43.29 | 4.22 | 24.62 | 72.13 |
| Babiche | 54.60 | 2.90 | 15.11 | 72.61 |
| Cabeche | 50.89 | 4.32 | 19.61 | 74.82 |
| Dacrons | 54.75 | - | 20.22 | 74.97 |
| Echevin | 45.18 | 1.53 | 30.61 | 77.32 |
| Faubers | 54.44 | - | 13.2 | 67.64 |
| Gargote | 48.22 | - | 20.34 | 68.56 |
| Hauynes | 45.58 | 5.58 | 14.99 | 66.15 |
| Inconel | 45.69 | 1.78 | 19.96 | 67.43 |
| Jobelin | 51.53 | - | 15.36 | 66.89 |
| Kochias | 52.49 | 10.56 | 8.91 | 71.96 |
| Labourg | 56.68 | - | 17.82 | 74.50 |
| Orillon | 51.89 | - | 13.42 | 65.31 |
| Generic | 58.60 | 1.83 | 22.04 | 82.47 |

According to the Realist Hypothesis, we would expect different judges to display different reasoning patterns. Indeed, feature importance varied wildly within a general trend. Cluster 1 features are clearly dominant across all models, reaching 58% for the generalist model. For 9 models out of 14, parents' petitions and child expressed preferences are above 50%. For 5 models out of 14, cluster 1 features are below 50%. Cluster 2 features are barely significant or not even registering in the top 15 features, except for the Kochias model. The Kochias model is the only one where Parental Behavior and parenting issues reached 10%. Cluster 3 features are significant but range from 8% to 30%. We observe that for 8 models out of 14, practical considerations are above 15%. For 6 models out of 14, practical considerations are below or equal to 15%.

## 9. Ablation study

**Table 24. F1 scores after ablation.**

| Bucket | Base RF | Ablation1 Petitions | Ablation2 Behavior | Ablation3 Practical | Base XGB | Ablation1 Petitions | Ablation2 Behavior | Ablation3 Practical |
|---|---|---|---|---|---|---|---|---|
| Anatole | 0.79 | 0.68 | 0.79 | 0.68 | 0.81 | 0.67 | 0.84 | 0.68 |
| Babiche | 0.78 | 0.61 | 0.78 | 0.75 | 0.83 | 0.6 | 0.81 | 0.77 |
| Cabeche | 0.77 | 0.66 | 0.79 | 0.64 | 0.77 | 0.65 | 0.78 | 0.77 |
| Dacrons | 0.79 | 0.68 | 0.81 | 0.71 | 0.78 | 0.65 | 0.76 | 0.71 |
| Echevin | 0.83 | 0.63 | 0.82 | 0.77 | 0.81 | 0.62 | 0.86 | 0.77 |
| Faubers | 0.81 | 0.51 | 0.81 | 0.71 | 0.79 | 0.51 | 0.76 | 0.68 |
| Gargote | 0.86 | 0.68 | 0.86 | 0.72 | 0.87 | 0.66 | 0.85 | 0.73 |
| Generic | 0.79 | 0.61 | 0.8 | 0.74 | 0.8 | 0.62 | 0.8 | 0.74 |
| Hauynes | 0.76 | 0.63 | 0.81 | 0.74 | 0.72 | 0.62 | 0.8 | 0.71 |
| Inconel | 0.85 | 0.69 | 0.81 | 0.65 | 0.82 | 0.66 | 0.82 | 0.62 |
| Jobelin | 0.81 | 0.51 | 0.8 | 0.78 | 0.78 | 0.5 | 0.77 | 0.83 |
| Kochias | 0.86 | 0.69 | 0.85 | 0.84 | 0.83 | 0.63 | 0.82 | 0.81 |
| Labourg | 0.82 | 0.72 | 0.88 | 0.84 | 0.81 | 0.69 | 0.81 | 0.76 |
| Orillon | 0.87 | 0.6 | 0.85 | 0.88 | 0.82 | 0.57 | 0.85 | 0.82 |

The ablation study further substantiates the centrality of Cluster 1 features (parental petitions and child preferences) in predicting appellate outcomes in custody disputes. Across both Random Forest and XGBoost models, removing these features (Ablation 1) leads to a consistent and often substantial decline in model performance, with accuracy drops reaching up to 0.30 in some cases (e.g., Faubers and Orillon). This performance degradation confirms that Cluster 1 features are not only statistically dominant in feature importance rankings but also functionally indispensable to the predictive capacity of the models.

In contrast, the removal of behavioral and parenting-related features (Ablation 2) produces minimal or even negligible effects on accuracy, reinforcing the earlier observation that Cluster 2 plays a limited role in distinguishing outcomes, despite its theoretical legal salience.

The ablation of practical features (Ablation 3) yields more heterogeneous results, with some models experiencing moderate accuracy declines (e.g., Echevin, Labourg), while others remain largely unaffected. This variability suggests that while practical considerations have situational relevance, their predictive utility is secondary, case-dependent and weighted quite differently among judges: some judges seems to take particular consideration of logistical imperatives and child disruption of daily routine, while others seem to place more weight in other factors. Overall, the ablation findings emphasize a critical disjunction between the normative weight that behavioral evidence might be expected to carry in custody determinations and the empirical patterns captured by machine learning models.

## References

[1] WU Yifan, Reducing Judicial Inconsistency through AI: A Review of Legal Judgement Prediction Models, ITM Web of Conferences 70, 02009 (2025), DAI 2024: doi.org/10.1051/itmconf/20257002009

[2] PAILLET, Guillaume, "Un tiers de divorces en moins en 15 ans", Infos Rapides Justice, Numéro 19, Service de la statistique, des études et de la recherche (SSER), 28 novembre 2024: <https://www.justice.gouv.fr/sites/default/files/2024-11/Infos_rapides_justice_n19_1.pdf>

[3] KOSTER, Tara and POORTMAN, Anne-Rigt. Fairness perceptions of the postdivorce division of childcare and child-related expenses. Journal of Marriage and Family, 2025, vol. 87, no 2, p. 751-771.

[4] MACKAAY, E., & ROBILLARD, P. (1974). Predicting judicial decisions: The nearest neighbour rule and visual representation of case patterns. University of Montreal, Law and Economics Research Paper.

[5] ZELEZNIKOW, J., & HUNTER, D. (1994). Building intelligent legal information systems: Representation and reasoning in law. Kluwer Law and Taxation Publishers.

[6] BRÜNINGHAUS, S., & ASHLEY, K. D. (2003). Combining case-based and model-based reasoning for predicting the outcome of legal cases. International Conference on Case-Based Reasoning.

[7] RINGGER, Eric K., et al. (2008). Assessing the costs of machine-assisted corpus annotation through a user study. In Proceedings of the 6th International Conference on Language Resources and Evaluation (LREC 2008).

[8] CHALKIDIS, Ilias, JANA, Abhik, HARTUNG, Dirk, et al. LexGLUE: A benchmark dataset for legal language understanding in English. arXiv preprint arXiv:2110.00976, 2021.

[9] ZHONG, Haoxi, XIAO, Chaojun, TU, Cunchao, et al. How does NLP benefit legal system: A summary of legal artificial intelligence. arXiv preprint arXiv:2004.12158, 2020.

[10] SHI, Zhiyi, KIM, Junsik, JEONG, Davin, et al. Surprisingly Simple: Large Language Models are Zero-Shot Feature Extractors for Tabular and Text Data. Under review as a conference paper at ICLR 2025.

[11] ZIN, May Myo, SATOH, Ken, et BORGES, Georg. Leveraging LLM for Identification and Extraction of Normative Statements. In : Legal Knowledge and Information Systems. IOS Press, 2024. p. 215-225.

[12] CAO, Lang, WANG, Zifeng, XIAO, Cao, et al. PILOT: Legal Case Outcome Prediction with Case Law. In : Proceedings of the 2024 Conference of the North American Chapter of the Association for Computational Linguistics: Human Language Technologies (Volume 1: Long Papers). 2024. p. 609-621.

[13] RIBEIRO DE FARIA, Joana, XIE, Huiyuan, et STEFFEK, Felix. Information extraction from employment tribunal judgments using a large language model. Artificial Intelligence and Law, 2025, p. 1-22.

[14] OUYANG, Long, WU, Jeffrey, JIANG, Xu, et al. Training language models to follow instructions with human feedback. Advances in neural information processing systems, 2022, vol. 35, p. 27730-27744.

[15] SONG, Meng, WANG, Xuezhi, BIRADAR, Tanay, et al. A minimalist prompt for zero-shot policy learning. arXiv preprint arXiv:2405.06063, 2024.

[16] AGARWAL, Bhavik, JOSHI, Ishan, et ROJKOVA, Viktoria. Think inside the json: Reinforcement strategy for strict llm schema adherence. arXiv preprint arXiv:2502.14905, 2025.

[17] ASHLEY, Kevin D. A brief history of the changing roles of case prediction in AI and law. Law Context: A Socio-Legal J., 2019, vol. 36, p. 93.

[18] ROSILI, Nur Aqilah Khadijah, ZAKARIA, Noor Hidayah, HASSAN, Rohayanti, et al. A systematic literature review of machine learning methods in predicting court decisions. IAES International Journal of Artificial Intelligence, 2021, vol. 10, n4, p. 1091.

[19] FENG, Yi; LI, Chuanyi; NG, Vincent. Legal Judgment Prediction: A Survey of the State of the Art. En IJCAI. 2022. p. 5461-5469.

[20] MEDVEDEVA, Masha, ÜSTÜN, Ahmet, XU, Xiao, et al. Automatic judgement forecasting for pending applications of the European Court of Human Rights. In : Proceedings of the Fifth Workshop on Automatec Semantic Analysis of Information in Legal Text (ASAIL 2021). CEUR Workshop Proceedings, 2021. p. 12-23.

[21] EPSTEIN, L., W. M. LANDES, and R. A. POSNER. “Inferring the Winning Party in the Supreme Court from the Pattern of Questioning at Oral Argument.” The Journal of Legal Studies. 2010. 39(2):433–467.

[22] KAUFMAN, Aaron Russell, KRAFT, Peter, et SEN, Maya. Improving supreme court forecasting using boosted decision trees. Political Analysis, 2019, vol. 27, no 3, p. 381-387.

[23] KATZ, Daniel Martin, BOMMARITO II, Michael J., et BLACKMAN, Josh. A general approach for predicting the behavior of the Supreme Court of the United States. PloS one, 2017, vol. 12, no 4, p. e0174698.

[24] RUGER, Theodore W., KIM, Pauline T., MARTIN, Andrew D., et al. The supreme court forecasting project: legal and political science approaches to predicting supreme court decisionmaking. Colum. L. Rev., 2004, vol. 104, p. 1150.

[25] KATZ, Daniel Martin, BOMMARITO II, Michael J., et BLACKMAN, Josh. Predicting the behavior of the supreme court of the united states: A general approach. arXiv preprint arXiv:1407.6333, 2014.

[26] SHARMA, R.D., MITTAL, S., TRIPATHI, S., ACHARYA, S. Using Modern Neural Networks to Predict the Decisions of Supreme Court of the United States with State-of-the-Art Accuracy. In: Arik, S., Huang, T., Lai, W., Liu, Q. (eds) Neural Information Processing. ICONIP 2015. Lecture Notes in Computer Science (2015). vol 9490. Springer, Cham.

[27] KATZ, Daniel Martin, BOMMARITO II, Michael J., et BLACKMAN, Josh. A general approach for predicting the behavior of the Supreme Court of the United States. PloS one, 2017, vol. 12, no 4, p. e0174698.

[28] KATZ, Daniel Martin, BOMMARITO II, Michael James, et BLACKMAN, Josh. Crowdsourcing accurately and robustly predicts Supreme Court decisions. arXiv preprint arXiv:1712.03846, 2017.

[29] KAUFMAN, Aaron Russell, KRAFT, Peter, et SEN, Maya. Improving supreme court forecasting using boosted decision trees. Political Analysis, 2019, vol. 27, no 3, p. 381-387.

[30] ALGHAZZAWI, Daniyal, BAMASAG, Omaimah, ALBESHRI, Aiiad, et al. Efficient prediction of court judgments using an LSTM+ CNN neural network model with an optimal feature set. Mathematics, 2022, vol. 10, no 5, p. 683.

[31] SULEA, Octavia-Maria, Marcos ZAMPIERI, Shervin MALMASI, Mihaela VELA, Liviu P. DINU, and Josef VAN GENABITH. Exploring the use of text classification in the legal domain. In Proceedings of ASAIL@ICAIL, volume 2143, 2017.

[32] SULEA, Octavia-Maria, Marcos ZAMPIERI, Shervin MALMASI, Mihaela VELA, Liviu P. DINU, and Josef VAN GENABITH. Predicting the law area and decisions of French supreme court cases. In Proceedings of RANLP, pages 716–722, 2017.

[33] ALMUSLIM, Intisar et INKPEN, Diana. Legal judgment prediction for Canadian appeal cases. In : 2022 7th International Conference on Data Science and Machine Learning Applications (CDMA). IEEE, 2022. p. 163-168.

[34] STRICKSON, Benjamin et DE LA IGLESIA, Beatriz. Legal judgement prediction for UK courts. In : Proceedings of the 3rd International Conference on Information Science and Systems. 2020. p. 204-209.

[35] NIKLAUS, Joel; CHALKIDIS, Ilias; STÜRMER, Matthias. Swiss-judgment-prediction: A multilingual legal judgment prediction benchmark. arXiv preprint arXiv:2110.00806, 2021.

[36] MALIK, Vijit, SANJAY, Rishabh, NIGAM, Shubham Kumar, et al. ILDC for CJPE: Indian Legal Documents Corpus for Court Judgment Prediction and Explanation. In : Proceedings of the 59th Annual Meeting of the Association for Computational Linguistics and the 11th International Joint Conference on Natural Language Processing (Volume 1: Long Papers). 2021. p. 4046-4062.

[37] NIGAM, Shubham Kumar; DEROY, Aniket. Fact-based court judgment prediction. En Proceedings of the 15th Annual Meeting of the Forum for Information Retrieval Evaluation. 2023. p. 78-82.

[38] LAGE-FREITAS, André, ALLENDE-CID, Héctor, SANTANA, Orivaldo, et al. Predicting Brazilian court decisions. PeerJ Computer Science, 2022, vol. 8, p. e904.

[39] ALETRAS, Nikolaos, TSARAPATSANIS, Dimitrios, PREOŢIUC-PIETRO, Daniel, et al. Predicting judicial decisions of the European Court of Human Rights: A natural language processing perspective. PeerJ computer science, 2016, vol. 2, p. e93.

[40] VISENTIN, Andrea, NARDOTTO, Alessia, et O'SULLIVAN, Barry. Predicting judicial decisions: A statistically rigorous approach and a new ensemble classifier. In : 2019 IEEE 31st International Conference on Tools with Artificial Intelligence (ICTAI). IEEE, 2019. p. 1820-1824.

[41] MEDVEDEVA, Masha, VOLS, Michel, et WIELING, Martijn. Judicial decisions of the European Court of Human Rights: Looking into the crystal ball. In : Proceedings of the conference on empirical legal studies. 2018. p. 24.

[42] MEDVEDEVA, Masha, VOLS, Michel, et WIELING, Martijn. Using machine learning to predict decisions of the European Court of Human Rights. Artificial Intelligence and Law, 2020, vol. 28, no 2, p. 237-266.

[43] CHALKIDIS, Ilias, ANDROUTSOPOULOS, Ion, et ALETRAS, Nikolaos. Neural Legal Judgment Prediction in English. In : Proceedings of the 57th Annual Meeting of the Association for Computational Linguistics. Association for Computational Linguistics, 2019.

[44] XU, Shanshan, SANTOSH, T. Y. S., ICHIM, Oana, et al. Through the lens of split vote: Exploring disagreement, difficulty and calibration in legal case outcome classification. arXiv preprint arXiv:2402.07214, 2024.

[45] CAO, Lang, WANG, Zifeng, XIAO, Cao, et al. PILOT: Legal Case Outcome Prediction with Case Law. In : Proceedings of the 2024 Conference of the North American Chapter of the Association for Computational Linguistics: Human Language Technologies (Volume 1: Long Papers). 2024. p. 609-621.

[46] XIAO, Chaojun, ZHONG, Haoxi, GUO, Zhipeng, et al. Cail2018: A large-scale legal dataset for judgment prediction. arXiv preprint arXiv:1807.02478, 2018.

[47] PENG Yi-Ting, LEI Chin-Laung. Using Bidirectional Encoder Representations from Transformers (BERT) to predict criminal charges and sentences from Taiwanese court judgments. PeerJ Comput Sci. 2024 Jan 31;10:e1841. doi: 10.7717/peerj-cs.1841. PMID: 38435559; PMCID: PMC10909178.

[48] LUO, Bingfeng, FENG, Yansong, XU, Jianbo, et al. Learning to Predict Charges for Criminal Cases with Legal Basis. In : Proceedings of the 2017 Conference on Empirical Methods in Natural Language Processing. Association for Computational Linguistics, 2017.

[49] ZHONG, Haoxi, GUO, Zhipeng, TU, Cunchao, et al. Legal judgment prediction via topological learning. In : Proceedings of the 2018 conference on empirical methods in natural language processing. 2018. p. 3540-3549.

[50] HU, Zikun, LI, Xiang, TU, Cunchao, et al. Few-shot charge prediction with discriminative legal attributes. In : Proceedings of the 27th international conference on computational linguistics. 2018. p. 487-498.

[51] WEI, Duan et LIN, Li. An external knowledge enhanced multi-label charge prediction approach with label number learning. arXiv preprint arXiv:1907.02205, 2019.

[52] YANG, Wenmian, JIA, Weijia, ZHOU, Xiaojie, et al. Legal judgment prediction via multi-perspective bi-feedback network. arXiv preprint arXiv:1905.03969, 2019.

[53] XU, Nuo, WANG, Pinghui, CHEN, Long, et al. Distinguish Confusing Law Articles for Legal Judgment Prediction. In : Proceedings of the 58th Annual Meeting of the Association for Computational Linguistics. 2020. p. 3086-3095.

[54] YUE, Linan, LIU, Qi, JIN, Binbin, et al. Neurjudge: A circumstance-aware neural framework for legal judgment prediction. In : Proceedings of the 44th international ACM SIGIR conference on research and development in information retrieval. 2021. p. 973-982.

[55] DONG Qian and Shuzi NIU. 2021. Legal Judgment Prediction via Relational Learning. In Proceedings of the 44th International ACM SIGIR Conference on Research and Development in Information Retrieval (SIGIR '21). Association for Computing Machinery, New York, NY, USA, 983–992. https://doi.org/10.1145/3404835.3462931

[56] LIU, Yifei, WU, Yiquan, ZHANG, Yating, et al. Ml-ljp: Multi-law aware legal judgment prediction. In : Proceedings of the 46th international ACM SIGIR conference on research and development in information retrieval. 2023. p. 1023-1034.

[57] ZHANG, Yue, TIAN, Zhiliang, ZHOU, Shicheng, et al. RLJP: Legal Judgment Prediction via First-Order Logic Rule-enhanced with Large Language Models. arXiv preprint arXiv:2505.21281, 2025.

[58] PENG YT, Lei CL. Using Bidirectional Encoder Representations from Transformers (BERT) to predict criminal charges and sentences from Taiwanese court judgments. PeerJ Comput Sci. 2024 Jan 31;10:e1841. doi: 10.7717/peerj-cs.1841. PMID: 38435559; PMCID: PMC10909178.

[59] VARGA, Dávid, SZOPLÁK, Zoltán, KRAJCI, Stanislav, et al. Analysis and prediction of legal judgements in the Slovak criminal. 2021. <https://ceur-ws.org/Vol-2962/paper13.pdf>

[60] WANG, Yuzhong, XIAO, Chaojun, MA, Shirong, et al. Equality before the law: legal judgment consistency analysis for fairness. arXiv preprint arXiv:2103.13868, 2021.

[61] WU Yifan, Reducing Judicial Inconsistency through AI: A Review of Legal Judgement Prediction Models, ITM Web of Conferences 70, 02009 (2025), DAI 2024: doi.org/10.1051/itmconf/20257002009

[62] MEDVEDEVA, Masha, VOLS, Michel, et WIELING, Martijn. Using machine learning to predict decisions of the European Court of Human Rights. Artificial Intelligence and Law, 2020, vol. 28, no 2, p. 237-266.

[63] BARALE Claire , Michael ROVATSOS, Nehal BHUTA, When Fairness Isn't Statistical: The Limits of Machine Learning in Evaluating Legal Reasoning, arXiv:2506.03913v1 [cs.CL] 04 Jun 2025

[64] MEDVEDEVA, Masha et MCBRIDE, Pauline. Legal judgment prediction: If you are going to do it, do it right. In : Proceedings of the Natural Legal Language Processing Workshop 2023. 2023. p. 73-84.

[65] FENG, Yi; LI, Chuanyi; NG, Vincent. Legal Judgment Prediction: A Survey of the State of the Art. En IJCAI. 2022. p. 5461-5469.

[66] LAGE-FREITAS, André, ALLENDE-CID, Héctor, SANTANA, Orivaldo, et al. Predicting Brazilian court decisions. PeerJ Computer Science, 2022, vol. 8, p. e904.